\documentclass[pdflatex,sn-apa]{sn-jnl}

\usepackage{graphicx}%
\usepackage{multirow}%
\usepackage{amsmath,amssymb,amsfonts}%
\usepackage{amsthm}%
\usepackage{mathrsfs}%
\usepackage[title]{appendix}%
\usepackage{xcolor}%
\usepackage{textcomp}%
\usepackage{manyfoot}%
\usepackage{booktabs}%
\usepackage{algorithm}%
\usepackage{algorithmicx}%
\usepackage{algpseudocode}%
\usepackage{listings}%
\usepackage[table]{xcolor}
\usepackage{hyperref}

\begin{document}

\title[Article Title]{Artificial Phantasia: Emergent Mental Imagery in Large Language Models}


\author*[1,2]{\fnm{Morgan} \sur{McCarty}}\email{research@morgannmccarty.com}

\author[2,3]{\fnm{Jorge} \sur{Morales}}\email{j.morales@northeastern.edu}

\affil*[1]{\orgdiv{Khoury College of Computer Sciences}, \orgname{Northeastern University}, \orgaddress{\street{360 Huntington Avenue}, \city{Boston}, \postcode{02115}, \state{Massachusetts}, \country{United States}}}

\affil[2]{\orgdiv{Department of Psychology}, \orgname{Northeastern University}}

\affil[3]{\orgdiv{Department of Philosophy}, \orgname{Northeastern University}}


\abstract{Can visual imagery be driven solely by language? This idea goes against cognitive science's traditional view that visual mental imagery is only possible through pictorial representations. Large Language Models (LLMs) provide nascent evidence not only that visual mental imagery via propositional-representations is possible, but that it can be more robust than human imagination. We created dozens of novel items for an extension to a classic task which is argued to be solvable exclusively via pictorial representations (i.e., language alone would be insufficient). Subjects were asked to imagine a series of compositional letter and shape transformations and identify the resultant ``image''. We found that the best LLMs performed significantly better than humans ($n = 100$ human participants, $p < .0001$), indicating the existence of an \textit{artificial phantasia}, or emergent ``visual'' mental imagery that may not be pictorial. Furthermore, we tested reasoning models with variable reasoning-token allocation and found that models perform best with longer reasoning chains, demonstrating a linguistic impact on the task---language alone may be sufficient. We examined three emergent imagery hypotheses: pure propositional imagery, propositional imagery with visio-linguistic priors, or pictorial visual imagery (classical visual imagery). Our study not only presents evidence for a previously unreported emergent cognitive capacity of LLMs, but also reignites debate on the requirement for a pictorial format in mental imagery.}




\maketitle

\section{Introduction}\label{sec:introduction}

Visual imagination---the mind's eye---is a core element of human existane and experience. Humans report seeing and manipulating ``images'' in this mind's eye and, collectively, cognitive science has named this phenomenon: ``mental imagery''. The idea of images being separable from the ability to complete ``imagery'' tasks is not new \citep{pylyshyn2002}, but generally it has been accepted that mental imagery operates on imagistic representations. But what if there was an alternate model of cognition which challenged this idea? Enter the Large Language Model (LLM).

LLMs have stormed ahead with new capabilities, features, and world impacts. Debate on these capabilities, however, primarily focuses on benchmark performance (e.g., \cite{kocisky2018, rein2023, hendrycks2021, wang2024, khan2023}). Slowly, but surely, however, researches have begun to implore LLMs as candidate subjects for cognitive science and psychological research. For instance, more and more recent work has targeted areas like introspection \citep{binder2025, lindsey2026, plunkett2025, ackerman2026} and theory of mind \citep{hu2023, kosinski2024, prakash2026}. Additionally, some benchmarks have been created which focus on spatial and visual reasoning \citep{chollet2025, ma2025}. In addition to cognitive exploration of LLMs as candidate subjects for experimentation, there is a wealth of evidence than neural networks, in general, are useful tools for understanding \textit{human} cognition \citep{demszky2023, mcgrath2024, leshinskaya2025, frank2025}, in particular visual perception \citep{chen2025, yamins2014}. With specific regard to visual mental imagery, a small amount of work has been done with benchmarking \citep{sepehri2026}, attempting to introduce mental imagery \citep{yang2025}, and using mental imagery as an evaluative framework for reasoning \citep{zeller2026}, but no prior work has explored the central questions of ``does this capability exist nascently?'' and, if so, ``is this capability analogous to humans, or does it provide an alternative model?''

\begin{figure*}[t]
\centering
\includegraphics[width=0.95\textwidth]{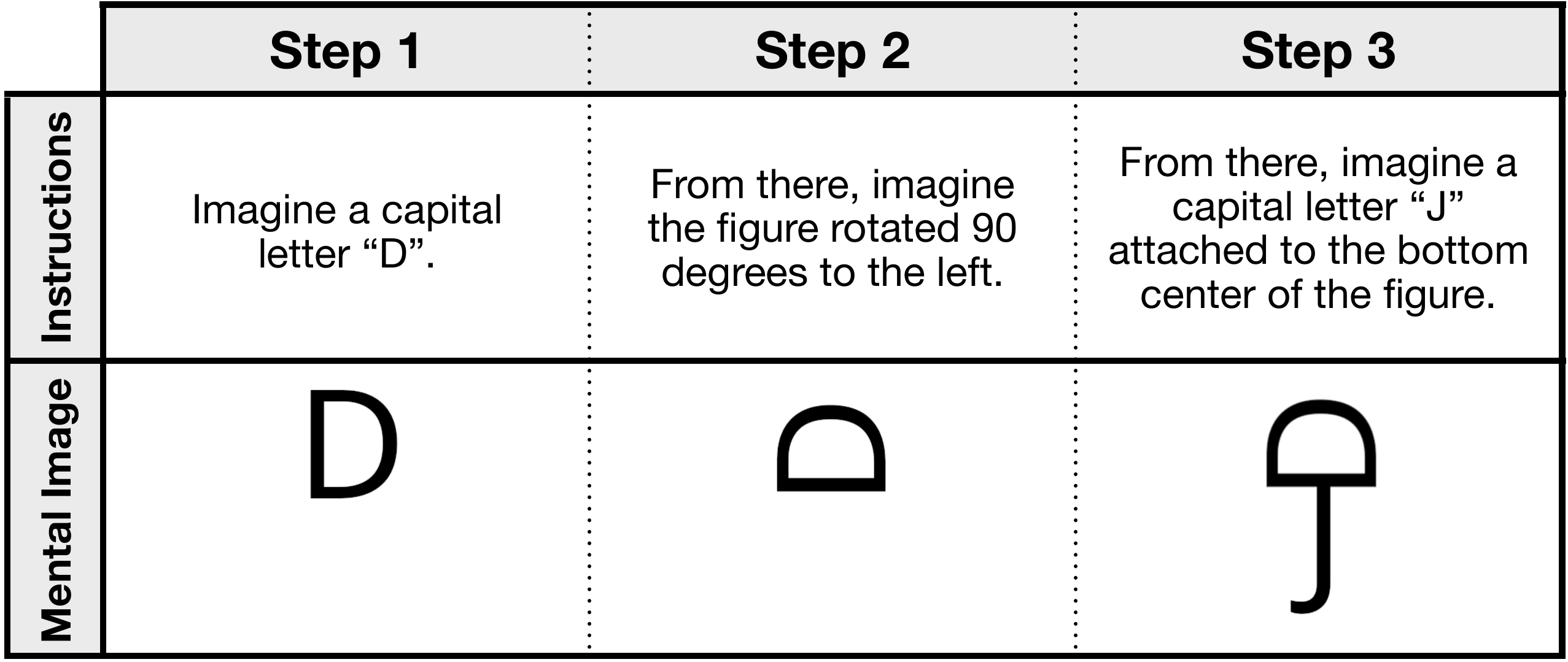}
\caption{One of the instruction sets introduced in \cite{finke1989}. Here, subjects are meant to recognize from the resulting mental image that the final imagined object looks like an \textit{umbrella}. The instructions have been rewritten to be clearer both for prompting LLMs, as well as for human understandability.}
\label{fig:umbrella}
\end{figure*}

Here we present a mental imagery task adapted from human research in cognitive psychology \citep{finke1989} which offers an opportunity for assessing models in novel and sophisticated ways. Subjects (models or humans) are presented with a series of instructions asking them to imagine a shade (usually a letter) or a transformation of the existing ``image''. The classic example (with revised instructions) is presented in Figure \ref{fig:umbrella}. We designed a series of extended stimuli sets which cannot exist in the training data of the tested models. Our data leaves great room for future model performance increases while simultaneously showing the broad diversity of existing models.

\section{Three Explanations for Emergent Visual Mental Imagery}

Leading corporate labs developing state-of-the-art models tend not to disclose their architectural design decisions, but from the best open-source models we can discover several key elements which may impact mental imagery. 

State-of-the-art models are large transformers \citep{vaswani2017} trained on vast swaths of the internet as their corpus \citep{qwen3tech, gpt5tech, geminitech}. These models can exceed many billions to trillions of parameters. Additionally, Vision-Language Models, like Qwen 3 VL, are often trained as text generation models with multimodal input \citep{ghosh2025}. This involves training an image encoder and aligning it to a pretrained language generation model such that visual inputs are encoded to the same initial embedding space of the language decoder (see \cite{qwen25tech} for a practical example of this implementation). Both the encoder and decoder are then unfrozen and training continues, often with text and images interspersed. From this paradigm, we might expect three variants of an emergent visual mental imagery to occur. Notably, when we refer to ``visual'' in this manuscript, we are referring to perceptual processing associated with vision, not necessarily the processing of pictorial representations \citep{kaski2002}.

\subsection{Pure Propositional ``Visual'' Mental Imagery}

In models which have no imagistic training, success on the visual imagery reconstruction task \citep{finke1989} indicates the usage of a mental imagery system built upon propositional reasoning \citep{pylyshyn1973, pylyshyn2002}. No primary visual information (i.e., information adapted from images) exists within the model due to a lack of visual information during training. All information about images would have to be derived from human (or possibly other AI model; \cite{sun2025}) generated text.

\subsection{Propositonal ``Visual'' Mental Imagery with Visio-Linguistic Priors}

In models which have imagistic training akin to \cite{ghosh2025}'s Text Generation with Multimodal Input, we may expect visual information which is not necessarily accessible textually in the training corpus to be ingrained into the model in a linguistic or pseudo-linguistic format---perhaps like Fodor's ``language of thought'': \cite{fodor1975}. If the underlying format is that of propositional reasoning, we could expect shortcuts to the compositional nature of our task to be found \citep{khandelwal2025} which may explain how models perform so well on our task despite the general lack of ability for models to compositionally generate \citep{ismayilzada2025}. Notably, recent evidence has shown that pictorial information is lost very early in the language decoder, and transferred to linguistic representation, even with image input \citep{tartaglini2025}.

\subsection{Visual (Pictorial) Mental Imagery}

With the resolution of the visual mental imagery format debate largely ruling that visual mental imagery in most humans is primarily driven by pictorial representations \citep{pearson2015}, we would be amiss to not consider the possibility that Large Language Models, particularly those with imagistic training, can solve imagery-dependent tasks through imagistic representations (i.e., manipulation and generation of pictures in a ``mind's eye'' analogue). However, this would require successful compositional image generation to explain success in our task, an ability in which models consistently show error \citep{vatsa2025}. While it is possible that models can generate internal images via shortcuts (like generating an umbrella after determining that the letters correspond to one), this would first require using a propositional representation and therefore would better fit under the ``Mental Imagery with Visio-Linguistic Priors'' umbrella.

\section{Solving Mental Imagery Tasks without Mental Imagery?}

Many tasks in everyday life involve the usage of mental imagery to some degree (e.g., navigation, planning and decision-making, mental simulation, episodic memory, organization, spatial reasoning, emotional engagement and regulation, among others; \cite{bocchi2016,shepard1971, palombo2018, wheeler2000, byrne2007, holmes2010, krasich2024}). Unsurprisingly, most humans report having conscious mental imagery. However, a small percentage (1-4\%) of the population---aphantasics---report no conscious mental imagery \citep{wright2024, faw2009, zeman2015, dance2022, zeman2024}. If the pictorial view were correct, we would predict aphantasics to be incapable of performing mental imagery tasks at all; but this is not what we find \citep{blomkvist2023, kay2024, pounder2022, Bainbridge2021}. Aphantasics perform (almost) at the same level as people with imagery. While there is a possibility that they rely on unconscious visual mental imagery \citep{nanay2021, michel2025}, aphantasics tend to report that they use verbal strategies \citep{keogh2021, kay2024}, giving renewed credence to the possibility of a purely propositional mental imagery.

In the human mind, language and imagery are deeply intertwined. It can be hard to evaluate the introspective reports of subjects (aphantasic or not) about the strategies that they use, as humans in general do not have good access to the inner processes that support their behavior \citep{nisbett1977}; or they may confabulate about the actual contents of their mental images \citep{bigelow2023}. State-of-the-art artificial systems such as LLMs offer a unique opportunity to test a system whose architecture and processing is primarily propositional, specifically, linguistic. Does this mean that there are types of reasoning that are just not available to them? Or is it possible that LLMs rely solely on their language-trained and language-processing architecture to achieve similar goals as humans who experience mental images? After all, as mentioned earlier, aphantasics are reported to perform at the same levels as imagers in a plethora of tasks previously thought to require mental imagery. The imagery debate does not seem, after all, to have been completely settled.\footnote{Aphantasia research has recently re-opened questions regarding the nature of mental imagery representations \citep{lorenzatti2025, lebon2025}. Proposals have ranged from unconscious pictorial representations \citep{michel2025, nanay2021} and absent pictorial representations to a preserved spatial imagery despite a diminished or absent object imagery \citep{Bainbridge2021, phillips2025}. The door for non-pictorial mental imagery has certainly reopened.} 

\subsection{Motivation for LLM Mental Imagery Tasks}
To test whether LLMs are capable of solving tasks designed to probe visual mental imagery, we gave several state-of-the-art models (Claude, Gemini, OpenAI), as well as several open-weight models (DeepSeek, Qwen, GPT oss), expanded, bespoke instruction sets following \citet{finke1989}'s approach described above. We also asked models with image capabilities to generate images in each step and to consider them in their answers. As the object reconstruction task we used is compositional (different images or aspects of images from each step need to be combined in subsequent steps to obtain the final answer), we conjectured that, direct, image-aided reasoning could increase performance, especially if the pictorial imagery framework is correct, and if forcing the models to produce and consider images in the intermediate steps could alter their approach to the task. We were especially interested in this as it forces the models to use their image encoder (something that does not occur during pure textual prompting as no pixels are input into the model). Finally, we obtained a human baseline for for this task by testing 100 human subjects. 

Whether LLMs can perform at a human level on this task is of intrinsic interest to understand what these new models can achieve. This type of object reconstruction is an ideal challenge for LLMs. The task is structured entirely through natural language (both the input and the output); the results are easily evaluated; and we included newly created examples that could not possibly be in their training set. Moreover, due to the nature of the task and the fact that it can be expanded, a human baseline can be straightforwardly established at any point. Additionally, LLMs performing at or beyond the human baseline would constitute evidence for LLM propositional reasoning-based imagery. The results from this task are also of interest for the cognitive science debate about the format(s) of mental imagery, since it would put to a test the idea that mental imagery necessarily involves some pictorial component.

\section{Results}\label{sec:results}

\subsection{Human Performance}

 Humans subjects exhibited an mean performance of 2.85 (out of 5: $\sigma = 1.51$) in our 60-item task. We established that all of the 48 new instruction sets were doable (though the hardest ones only received one or two meaningful answers from the entire subject population). Subjects received close to perfect scores on the easiest instruction sets, with only one or two non-meaningful answers. This offered us a good range of difficulties which was optimal for testing both human and artificial model abilities. (See Appendix \ref{sec:difficulty} for more information on item difficulty estimation and difficulty distribution).


Subjects' VVIQ scores were well within expected bounds and we found a small negative correlation between performance and mental imagery capacity VVIQ (see Appendix \ref{sec:vviq} for further discussion of VVIQ results).

\subsection{Recent Rapid Advancement}

\begin{figure*}[t]
\centering
\includegraphics[width=0.95\textwidth]{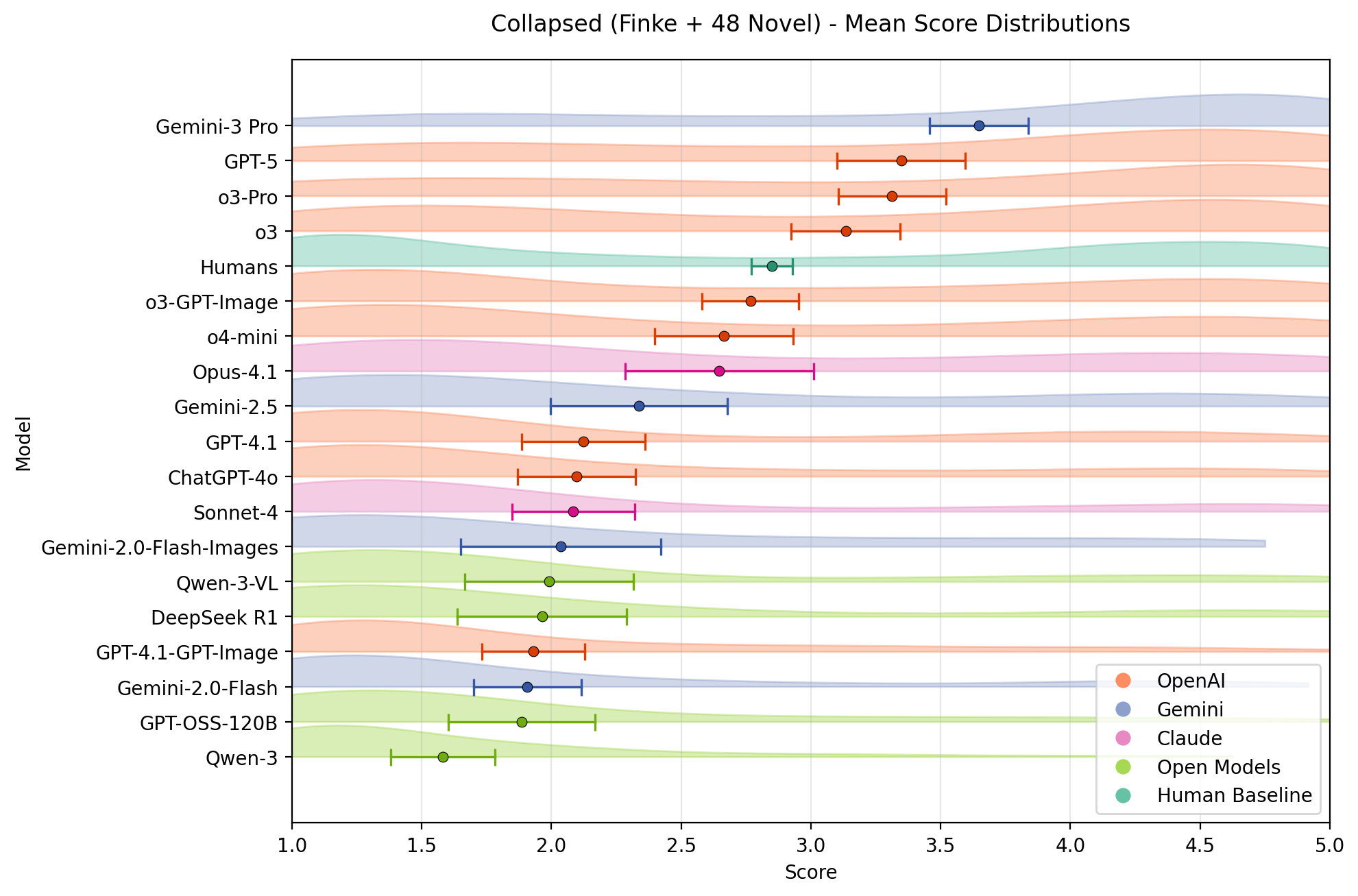}
\caption{Performance results in humans and LLMs. Data shows mean of score distributions for all tested models as well as KDE distributions over received scores. Only Gemini 3 Pro, GPT-5, and the o3 family significantly surpass the human baseline. Error bars indicate 95\% confidence intervals.}
\label{fig:collapsed_ci}
\end{figure*}

OpenAI's o3 model family, as well as GPT-5, vastly outperformed every preexisting model, surpassing the human baseline (comparisons against humans' mean score of $2.85$; o3 mean $=3.13$, $\sigma=1.44$, Mann-Whitney U Statistic $=110882$, $p=.0067$; o3-Pro mean $=3.31$, $\sigma=1.42$, U Statistic $=103354$, $p<.0001$; GPT-5 mean $=3.35$, $\sigma=1.38$, U Statistic $=67937$, $p=.0004$). Furthermore, Gemini 3 Pro reached a new frontier on our task: demolishing the human baseline with $3.65$ mean score ($\sigma=1.31$, U Statistic $=85943$, $p<.0001$).\footnote{Statistical significance results comparing humans to models was determined after correcting for multiple comparisons using Bonferroni: $\alpha = \frac{0.05}{18} = 0.0028$.} We show the stark gap between o3, GPT-5, Gemini 3 Pro, the human baseline, and every other model we tested in Figure \ref{fig:collapsed_ci}. Based on our scoring methodology (Appendix \ref{sec:grade_weight}), we graded the outputs of each model (and the human baseline) and calculated the mean score.

The only other models to perform non-significantly different from the human baseline on our task were o4-mini (mean score: $2.66$, $\sigma=1.48$, U Statistic $=87204$, $p=.441$) and Claude Opus 4.1 (mean score: $2.65$, $\sigma=1.43$, U Statistic $=43910$, $p=.590$). From a na\"ive perspective, we were surprised that adding external image generations to o3 (via GPT-image-1) significantly decreased the model's performance (mean score: $2.77$, $\sigma=1.47$) although it was still comparable to the human baseline (o3 with GPT-image-1 vs. o3 without GPT-image-1: U Statistic; humans vs o3 with GPT-image-1: U Statistic: $171586$, $p=.673$). The older or cheaper models, however, performed very poorly across the board. We looked at two other models from OpenAI: ChatGPT-4o and GPT-4.1 (with and without images), both performed significantly worse than the o3 model family. Claude Sonnet, Gemini 2.5 Pro, and Gemini 2.0 Flash all performed poorly. Gemini 2.0 Flash with images performed the worst of all proprietary models we tested.

Our testing of open models was extremely disappointing with all models performing worse than most, if not all, of the proprietary models. We discuss potential reasons for why this may be the case in Section \ref{sec:open_model_failure}.

\subsection{Reasoning Variation} \label{sec:reasoning}
\begin{figure*}[t]
    \centering
    \includegraphics[width=0.95\linewidth]{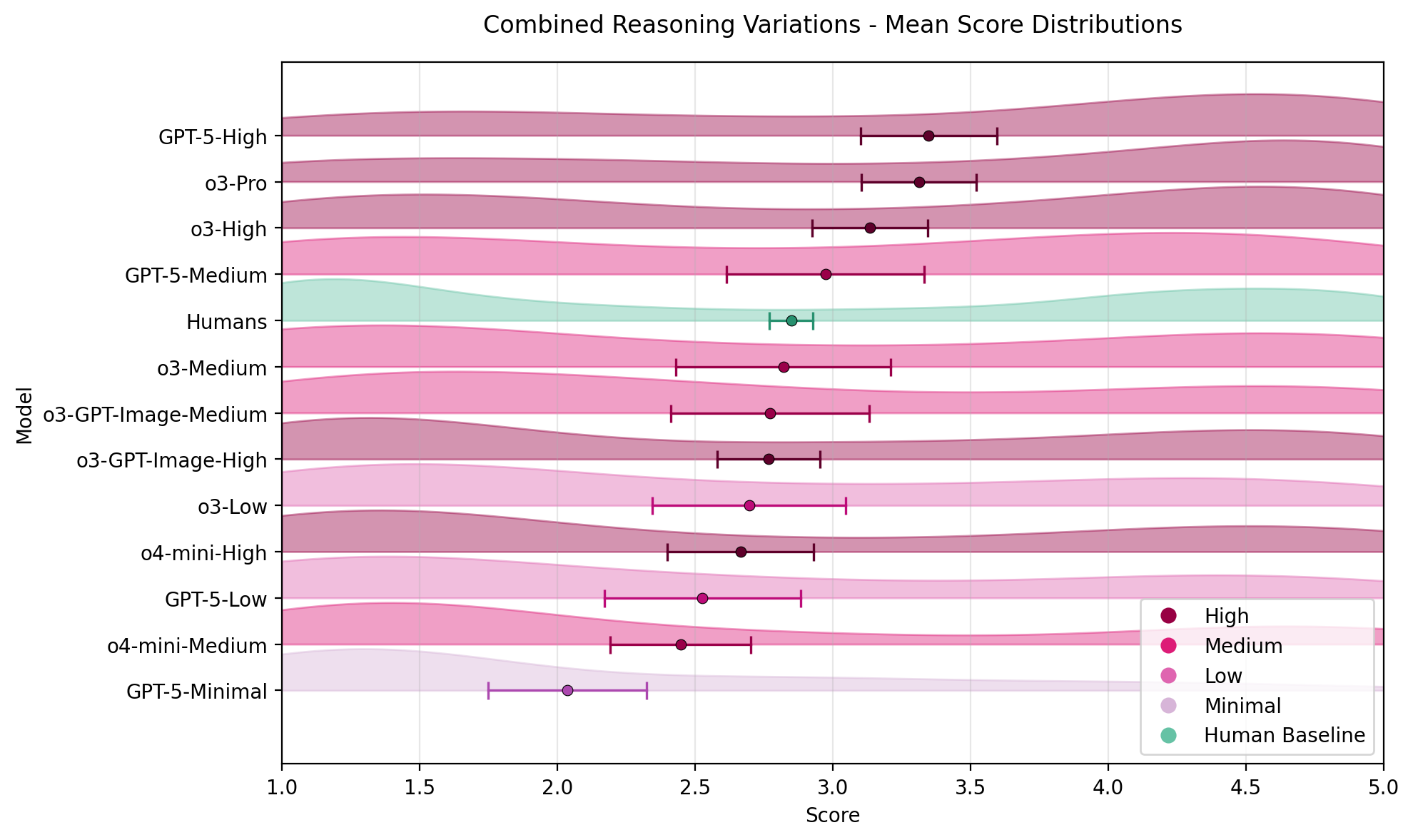}
    \caption{Reasoning parameter token modification results. More saturated colors indicate more reasoning-tokens allocated. Means and 95$\%$ confidence intervals shown overlayed on KDE distributions.}
    \label{fig:reasoning_collapsed_ci}
\end{figure*}

We also measured the effect of the `reasoning\_effort' parameter on the best performing OpenAI models. Across the board we saw clear evidence that more reasoning-tokens improved results. GPT-5 with `minimal' reasoning saw a performance drop to $2.04$ mean score ($\sigma=1.14$, from $3.35$). `Medium' and `low' reasoning saw similar shifts in-between the `minimal' and `high'. We display the results of this ablation experiment in Figure \ref{fig:reasoning_collapsed_ci}. The difference in GPT-5 between ``high'' reasoning and other reasoning levels\footnote{Means and SDs. ``high'': mean $=3.35$, $\sigma=1.38$; ``medium'': mean $=2.97$, $\sigma=1.42$; ``low'': mean $=2.53$, $\sigma=1.41$; ``minimal'': mean $=2.04$, $\sigma=1.14$} was strongly significant for both ``high'' to ``low'' and ``high'' to ``minimal'' (``high'' vs. ``low'': U Statistic $=4778$, $p=.0003$; ``high'' vs. ``minimal'': U Statistic $=5482$, $p<.0001$)\footnote{Using Bonferroni for 7 comparisons: $\alpha=\frac{0.05}{7}=0.0071$}. GPT-5 ``high'' and ``medium'' did not show significant differences (U Statistic $=4202$, $p=.0676$). One interesting data point, however, was the lack of any significant difference between `high' and `medium' in o3 with GPT-image-1\footnote{Means and SDs. ``high'': mean $=2.77$, $\sigma=1.47$; ``medium'': mean $=2.77$, $\sigma=1.41$}. This may hint at the possibility that the image paradigm has, despite models overall lower performance, the potential to provide some support solving this task. The issues with compositional generation inherent to image generation models, however, may hinder it overall \citep{huang2023}.

\subsection{Uncertainty} \label{sec:idk}

One remarkable result we found was the lack of capability for LLMs to answer with uncertainty (and yet, it was unsurprising given well-known limitations of LLMs). We noted 54 occurrences of humans responding some form of ``I don't know'' to an instruction set, and 0 occurrences in the LLM responses. It is highly possible that LLMs provide responses even when great uncertainty exists (especially in our hardest instruction sets). This is an area worth exploring in further analyses on how LLMs succeed or fail at the task \citep{brahman2024, kadavath2022, malinin2021}.

Interestingly, when asked to report the imagined object when given no instructions, open models reported answers like ``blank slate'' or ``artificial neuron'' implying some ability to give ambiguous responses in cases of uncertainty. The impact of this on our task needs to be explored further by future researchers.

\subsection{Image-aided Generation}

Generating images, and solving the task using them, produced disappointing results. Perhaps unsurprisingly, in all cases of our image-aided paradigm the model performance dropped or, at best, stayed the same\footnote{Bonferoni $\alpha = \frac{0.05}{3} = 0.0167$.}\footnote{o3 with GPT-image-1: mean $=2.77$, $\sigma=1.47$; GPT-4.1: mean $=2.12$, $\sigma=1.32$; GPT-4.1 with GPT-image-1: mean $=1.93$, $\sigma=1.10$; Gemini 2.0 Flash: mean $=1.91$, $\sigma=1.16$; Gemini 2.0 Flash with images: mean $=2.04$, $\sigma=1.21$}: o3 vs. o3 with GPT-image-1 (U Statistic $=24828$, $p=.0087$); GPT-4.1 vs. GPT-4.1 with GPT-image-1 (U Statistic $=7223$, $p=.788$); Gemini 2.0 Flash vs. Gemini 2.0 Flash with images (U Statistic $=2170$, $p=.711$). We note, however, that o3's strong performance allowed greater room for a significant decrease. It is unclear what the overall effect of image-aided reasoning is, as the models still found some success (though diminished), and more exploration of its effects is needed \citep{yang2025, wu2024}. Notably, modification of the `reasoning\_effort' hyperparameter led to almost identical results, unlike what happened with the standard models without images (See Section \ref{sec:reasoning}).

\subsection{Multiple- vs. Single-Context}

We initially ran each model (excluding Gemini 2.0 Flash with images and o3 with images due to token limitations) in both a single context for all instructions as well as in a new context for each set of instructions (multiple-context). In our analysis of o3 and o3-Pro we found that including previous examples in context did not significantly change the overall performance (o3 SC: mean $=3.18$, $\sigma=1.35$ vs. o3 MC: mean $=3.11$, $\sigma=1.48$; Wilcoxon Signed-Rank Test Statistic $=442$, $p=.712$). This indicates that in-context learning was not significantly beneficial for this task. Because of this, we did not differentiate between context types for statistical analysis and instead analyzed results from both types of context together. Additionally, when we further tested several additional models with different reasoning levels, we only tested them in multiple-context. We graph the performance of the context interval variations of all tested models in Supplemental Figure \ref{fig:sc_mc_ci}.



\subsection{Open Model Failure}
\label{sec:open_model_failure}

Across the board, the open models we tested performed horribly. Qwen 3 was the single worst performing model, and no open model met or surpassed the human baseline. We are not entirely sure as to the cause of this (due to not having access to architectural specifications of proprietary models), but, nevertheless, we were able to learn at least one somewhat interesting point.

Qwen 3 (mean $=1.58$, $\sigma=0.787$) and Qwen 3 VL (mean $=1.99$, $\sigma=1.27$) did not have a significant difference in performance (U Statistic $=1443$, $p=.109$), but there was a trending difference in their means. We found this notable due to the architecture of the model \citep{qwen3tech}; most critically, the vision-language alignment step. By aligning visual information to a text embedding, the model learns how to represent visual information textually (and therefore propositionally). The exact impact of this step on our task is unclear, however, as the architecture of closed models is unknown.



\section{Discussion} \label{sec:discussion}

Our results were surprising. LLMs successfully accomplished a task in which humans are believed to rely (almost) exclusively on visual mental imagery. They also offer an interesting window into the capacities of advanced artificial language systems to provide insight on tasks that are thought to require something beyond what language models are capable of. Gemini Pro 3, GPT-5 and the o3 Language Reasoning Model family outperformed humans across the board. These models do not have any (known, immediate) access to any form of image processing built-in by default when dealing with text decoding. Additionally, we can infer that no image generation was being used unless explicitly called upon due to the decreased cost of simple inference. Thus, these models should not have been able to complete our task so successfully given the dominant pictorial views on mental imagery. It would seem, however, that these LLMs completed our task via language token manipulations. 

However, there is an alternative explanation, potentially orthogonal to the pictorial and propositional accounts. On this view, mental imagery is not a monolithic capacity; rather it must be distinguished into two capacities: spatial and object imagery \citep{phillips2025, teng2025}. According to this distinction, imagining objects and their surface features is supported by neural and psychological mechanisms that are different from the mechanisms supporting imagining the \textit{spatial relations} between objects. Evidence for this position comes from neurological patients with lesions resulting in preserved spatial reasoning abilities despite losing conscious mental imagery \citep{farah1988}. It is also supported by behavioral work with aphantasics who exhibit normal results in spatial imagery questionnaires and (almost) normal performance in mental imagery tasks that rely heavily on spatial relations such as remembering a scene and mental rotation, which is clearly related to the task we used here \citep{Bainbridge2021, Dawes2022}. See Appendix \ref{sec:vviq} for discussion of aphantasic subjects in our sample.

While LLMs' architecture, training, and data processing is primarily based on text, text-only models can acquire representations of spatial concepts \citep{patel2022mapping-85b} and develop internal representations matching a spatial layout solely through procedural descriptions of moving through spaces. Moreover, embeddings from different modalities (e.g., vision and text) can be mapped to a shared representational space \citep{radford2021, girdhar2023imagebind-566}. It is thus possible that while relying solely on textual token manipulation, the most advanced LLMs are still able to extract the required spatial relations of the objects in each step of our instruction sets to perform at or above human level. This interpretation may help bridge different theories of mental imagery.

Notably, there is some probability of the original 12 items in the Finke et al. task being in the training data. This, however, cannot easily explain away our results. First, only 12 out of 60 instruction sets appeared in their original paper; the rest of them are completely novel designs. Second, the novel items include rather idiosyncratic examples that would be astounding if anything like them were explicitly in the training data (e.g., a computer mouse made with an S, a C, a horizontal line, and a D). Third, and related to the previous point, even if the \textit{type} of task is present in the training data, and even if LLMs have information about the shape of letters (e.g., they know an 'o' is like a circle and they know ice cream can be represented with circles), the models still have to solve a challenging compositional task: they need to put together all the elements, in the right spatial layout, in some cases after 4 transformations including rotation and mirroring, and only then use the resulting composite novel shape to ``read out'' the actual answer. This demanding task goes well beyond knowing that ice cream icons have circles (notoriously, an enormous amount of other things can also be drawn using circles). Fourth, no model, not even the best performing ones, universally solved all the original Finke items, suggesting that the models are not simply finding a ready-made solution from their training data. Finally, in the case of some models (e.g., GPT 4.1 and o3) we have reasons to believe that they may share their training data (both have a June 01, 2024 knowledge cutoff). Even if some information relevant to our task were buried in the models' training data, it is the parameters and general architecture of the models that ultimately determine how well they were able to find relevant information and solve our novel items. Thus, our data shows that some advanced LLMs can reason about composite shapes and their spatial layout by relying solely on linguistic embeddings in ways that seriously challenge the hypothesis that a pictorial format is necessary for completing our task.

Frontier LLMs give a new perspective to the debate on the formats of mental imagery, re-opening questions about the necessity of iconic mental imagery and the adequacy of classic tasks to test it. Furthermore, our results also open an important question in computer science about the kind of tasks that our most advanced language models can accomplish providing an opportunity to create new, complex benchmarks in sophisticated cognitive tasks.

\section{Conclusions}

Advancements in Large Reasoning Models have progressed so quickly that testing all their possible emergent properties poses serious challenges. Our research shows how to test one such property: propositional reasoning-based mental imagery. We determined that Gemini Pro 3, GPT-5 and the o3 family of models is more than capable of solving problems traditionally thought to require visual imagery. Furthermore, we note the difference in capabilities between OpenAI, Claude, and Gemini models in this form of advanced reasoning. Our instruction sets were created \textit{ex novo}, so we can be confident that the performance we recorded is not an artifact of contaminated training data. Additionally, models did not universally succeed on the original items, which \textit{could} exist within their training data. Our results could be deeply impactful, both for the artificial intelligence community, for examining and identifying a capability not yet measured within the most advanced models, and also for the cognitive science community, for discovering results that provide insight into the strategies and techniques used to accomplish imagery-dependent tasks. Through further analysis and experimentation on Large Language Reasoning Models, we may discover more about different ways in which the human mind can work.

\newpage

\section{Methods}\label{sec:methods}

\subsection{Experimental Design}

\begin{figure*}[h]
\centering
\includegraphics[width=0.95\textwidth]{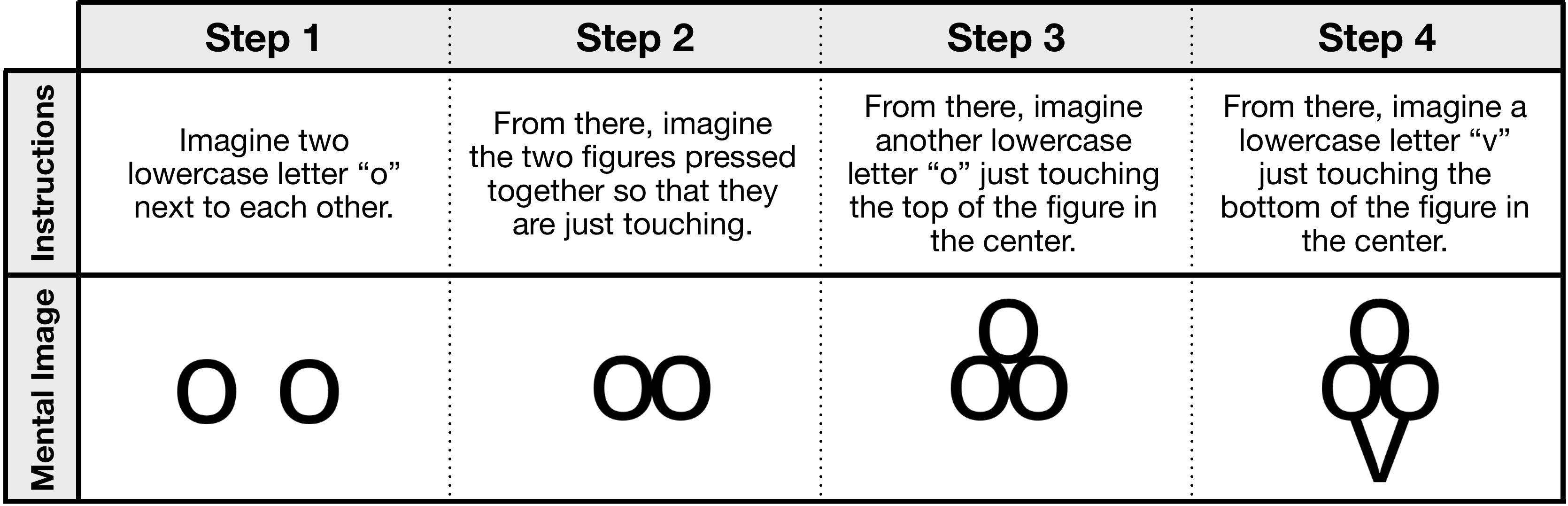}
\caption{One of our new instruction sets demonstrating the slightly increased cognitive complexity and more ambiguous canonical form (``balloons'', ``flower bouquet'', or ``ice cream'', among others). Note the usage of two letters in the first step, the abstract reference to existing symbols and scenes, and the final shape not being determinable until the final step.}
\label{fig:newtasks}
\end{figure*}

We used 60 instruction sets for an object recognition task. The set consisted of 48 completely novel examples, which we created specifically for this study, and the remaining 12 were taken from the original study by Finke et al. with modification (standardization of instructions and clarifications on ambiguities). Our new items varied in number of steps (2 to 4), final object, and overall complexity (Figure \ref{fig:newtasks}). We aimed to keep Finke et al.'s restriction that the final shape should not be identifiable until after the final transformation. Some of the tasks resulted in the same (or very similar) final shapes (e.g., three different ways of arriving at a shape that represented ``glasses'' or ``binoculars''), but as all combinations were unique and involved novel transformations, we believe this is not of concern.

Our new items integrated several changes to the original ones developed by Finke et al. Most notably, we allowed one step to include up to two letters, rather than just one, thus increasing cognitive load \citep{miller1956, farrington2011} but allowing more varied scenes. Additionally, we did not restrict the final image to having only one canonical form. The intended canonical form was not always immediately obvious, though at least one form was always clear. Our items' difficulty had a wider range, which we confirmed after establishing a new human baseline (see Appendix \ref{sec:human_baseline}). The complete instruction-sets are included in the project's anonymous GitHub repository (see Appendix \ref{sec:code_and_data}).

Finally, we updated the language in the 12-items taken directly from Finke et al. to match the format of our improved versions. Notably, ambiguous language was clarified (e.g., specification of capital versus lowercase letters), subsequent instructions were modified to reference earlier instructions only abstractly (e.g., ``the existing symbol'' versus ``the `E'''), and 180 degree rotations were changed to flips (when vertical) or mirroring (when horizontal). We did not ask models or human subjects to guess the resulting shape at each step, unlike the original experiment, only its final form.

We should highlight that because these 48 new items were created \textit{ex novo} for this experiment, it is highly improbable that any of them were present in the training data of any of the LLMs, and it is materially impossible that all of them were. These image reconstructions were of novel shapes in unique steps from anything in the extant literature. This is of crucial importance for testing the emerging reasoning capacities of these types of models.




\subsection{Image-Aided Instructions}

For each model that had a compatible image generation pipeline (excluding GPT-5\footnote{GPT-5 was released after we collected data on our image paradigm and found that it did not succeed for other models. As no further upgrades for GPT-image-1 were additionally released we chose not to test GPT-5 with GPT-image-1.}), we ran a modified version (Image-Aided) of the standard version (Language-Only) described above. In this modified image-aided version, models were prompted to generate images and modify those images (see Supplemental Figure \ref{fig:imageoutput} for an example), rather than imagine. Whenever possible, we kept reasoning enabled within the models. For Gemini we used the native image generation capabilities of 2.0 Flash's image generation preview version, and, for OpenAI, we used the native GPT-image-1 image generation tool integration. 

\subsection{Model Selection}

We gave the 60 instruction sets to three consumer accessible groups of models: Claude, Gemini, and OpenAI. For OpenAI reasoning models, `reasoning' was enabled and `reasoning\_effort' set to `high'. For older Gemini models, `thinking' was set to `dynamic'. For Gemini 3 Pro, `thinkinglevel' was set to `high'. For Claude Sonnet 4 we allocated 4000 tokens for extended thinking; for Claude Opus 4.1 we allocated 9000. For models that allow temperature modification, the value was set to 0.1 (we discuss some impacts of this in Appendix \ref{sec:temperature}). All other parameters were kept at default values. Our specific model choices are outlined in Supplemental Table \ref{tab:modelselection}.

For our initial analysis, we performed each experiment twice: once in a single context for all instruction sets (single-context), and once with a new context for each instruction set (multiple-context). This was done to see whether there was any significant difference in performance due to in-context learning \citep{dong2024, wurgaft2025}. The instruction sets were presented to each model in the same random order.

After completing our testing of the proprietary consumer models, we further tested four open-weight models: Qwen 3, Qwen 3 VL, DeepSeek R1, and gpt-oss-120b. We left all parameters to their default specifications, with exception of the `reasoning\_level' parameter for gpt-oss-120b which we set to `high' (open models were run through OpenRouter).

\subsection{Model Temperature} \label{sec:temperature}
In our initial testing, with the exception of Gemini 2.5 Pro (where we set the temperature to 0.1), all reasoning models were restricted to high temperatures by their respective APIs (1.0 for Claude and OpenAI models). After considering that the performance of Gemini 2.5 Pro (well below the other frontier reasoning models) may be due to the low temperature, we performed a new test using Gemini 3 Pro.

Temperature has been shown to meaningfully affect reasoning outputs \citep{wang2024Reasoning}, though on many tasks it does not necessarily improve the results \citep{renze2024}. We performed a hyperparameter search of the `temperature' parameter using values 0.1, 0.55, and 1.0 upon the release of Gemini 3 Pro. We were curious if modification in temperature would allow the model to explore more creative paths to the correct answer and succeed more often \citep{turing1950}. We found no significant difference between any of the temperature values tested.

\subsection{Human Baseline}
\label{sec:human_baseline}

We recruited 100 adult participants online. Each was given a random subset of 15 of the 60 instruction sets (Finke et al. sets and our 48-Item expansion set) for a total of 1500 submitted answers. Each instruction set had between 21-27 responses. The instruction sets were administered through a Qualtrics XM survey in a random order. To prevent textual biases from impacting the results (e.g., seeing a `d' on a screen affecting the specific shape of the imagined `d'), the instructions were played through audio recordings. Each occurrence of a letter in an instruction was modified to include the corresponding phonetic word from the International Radiotelephony Spelling Alphabet to increase the clarity of individual instructions (e.g., ``Imagine a B, as in Bravo.''). All audio was recorded by the same speaker (Author `Anonymous') and there was only one version of each audio instruction. If the same instruction appeared in two different sets, each set was given unique recordings. Participants were able to listen to the instructions as many times as needed, and were asked to imagine shapes without any visual aids (e.g., drawing). Finally, participants completed the Vividness of Visual Imagery Questionnaire (VVIQ) to assess their overall mental imagery capacity \citep{marks1973} (see Appendix \ref{sec:vviq}). Subjects participated for payment and were recruited through the online platform Prolific with an evenly distributed gender quota. To ensure intelligibility of the instructions, only subjects from the United States who reported English as their first language were sampled.

\subsection{Performance Evaluation}

Given the subjective nature of the task (e.g., Are ``balloons'', ``flower bouquet'', or ``ice cream'' all equally valid as an answer for the instructions in Figure \ref{fig:newtasks}?) and the potential for correct but variable answers, we recruited 376 na\"{i}ve subjects on Prolific to grade answers (LLM and human) from the studies described above. Subjects were shown a label along with a corresponding image approximating the final outcome of a set of instructions. They were asked to rate the reasonableness of the label describing the image we created \textit{in lieu} of target mental images. We collected 2030 unique answers from LLMs and humans in our mental imagery experiments. We excluded 122 of these answers because they were nonsensical, explicitly non-answers (e.g., ``unsure'', ``I don't know''; see Appendix \ref{sec:idk} for discussion), restatements of the instructions, or sexually explicit. We ended with a final set of 1911 valid answers to use as labels. Each subject grader was given 30 of these labels with the prompt: ``How well does this image represent $<$label$>$?'' and they were asked to respond on with a 5-point scale: ``Not at all'', ``A little'', ``Moderately'', ``A lot'', ``Completely''. In addition, the authors provided independent ``expert'' evaluation of all of the labels. The final score used to grade each valid response in our study was a weighted average of the expert's grades and the outsourced grades given by the Prolific subjects (see Appendix \ref{sec:grade_weight}).

\backmatter





\bmhead{Acknowledgments}

This work was funded, in part, by a Northeastern University Undergraduate Research and Fellowships PEAK Summit Award to M.M. The authors thank Dillon Plunkett for his feedback on the experimental design and Michael McPhee, Michaela Klimova, and Xueyi Huang for their feedback on the manuscript. Finally, we thank Glyn Mardis for her help in devising the stimulus set.





\bigskip





\begin{appendices}

\section{Access to Code and Data} \label{sec:code_and_data}

Our code and data is accessible under MIT License in our GitHub repository (\url{https://github.com/subjectivitylab/artificial_phantasia}). All human subject data has been de-identified. Please refer to the README.md file in the repository for usage and explanatory information regarding the repository. 

We provide code to run the experiment, view and re-perform our data analysis, and generate surveys for Qualtrics XM. Our dataset includes the instructions used for our task (along with the intended canonical form), the de-identified data from both humans and LLMs, the de-identified response ranking data, and the image representations of each intended image (the images from \citealp{finke1989} are their original versions for replicability).

\section{Extended Performance Evaluation - Humans}

\subsection{Grade Weighting} \label{sec:grade_weight}

In our initial work on grading the task we ran into two key issues: first, the issue of subjectivity as described previously; second, some answers were extremely literal and, as such, did not represent a new construction. Our initial plan was to grade similarly to \cite{finke1989}, but the subjectivity issue necessitated us exploring different options.

After recruiting our first set of subjects, we discovered a key issue with the responses. Namely, hyper-literal responses (e.g., `bd' after asking subjects to imagine a `b' next to a `d' and pressing them together) were graded very highly. We attempted to control for this by instructing subjects to rate such examples poorly, but, unfortunately, these were routinely graded highly. Our solution to this issue was to have the authors grade all 1911 responses in addition to the 376 na\"{a}ve subjects.

Almost all responses had 5 subject ratings, a few responses had 6, a very small number 7, and a single response 4 due to the random distribution methods of Qualtrics XM. In all cases we generated a `normal\_score' from the mean of all of the responses. In addition, we generated an `expert\_score' from the mean of the authors' ratings. Our final `overall\_score' was the average of the `normal\_score' and the `expert\_score'. All scores for all responses used in grading are available in our dataset (see Appendix \ref{sec:code_and_data}).

The subject ratings were distributed very heavily towards ``Not at all'' (or a grade of 1) due to most examples being difficult to justify (due to misconstructions, possible confusion in the case of humans, or possible hallucinations in the case of LLMs). The expert scores were distributed similarly, but with a noticeably higher occurrence of ``Completely'' (or a grade of 5). Knowledge of the creation of the items and their intended canonical result, along with a reduced bias towards the drawn representations of the intended mental image likely account for this. 

\subsection{Difficulty Analysis} \label{sec:difficulty}

Concurrently with the experimental task, we asked all participants to report how clear they found the instructions, as well as how identifiable they found the final mental image (both on a 5 point scale). These ratings as well as their standard deviations as a measure of consistency, along with the ratio of unique responses given to each answer, the number of instruction steps, and the number of imagined objects, were used to rank each of the instruction sets on the following weighted difficulty scale: 

\begin{center}
    \begin{align*}
    \text{Item Difficulty}     = \: & (0.20 \times \text{Total Instruction Steps}) \\
                               + \: & (0.20 \times \text{Total Objects}) \\
                               + \: & [0.15 \times (6 - \text{Mean Clarity Ratings})] \\
                               + \: & [0.15 \times (6 - \text{Mean Identifiability Ratings})] \\
                               + \: & (0.10 \times \text{Clarity Ratings Standard Deviation}) \\
                               + \: & (0.10 \times \text{Identifiability Ratings Standard Deviation}) \\
                               + \: & (0.10 \times \text{Unique Response Ratio}) \\
                               \\
    \text{Unique Response Ratio} = \: & \frac{\text{Unique Responses Per Label}}{\text{Total Responses Per Label}}  \\
    \end{align*}
\end{center}

Overall, we found that our set of instructions had a wide range of difficulty levels (see Supplemental Figure \ref{fig:diff_dist} for the distribution). We confirmed that the items that we designed to be harder (e.g., constructing a ``computer mouse'') indeed came out ranked as the most difficult, whereas the easiest ones (e.g., constructing a ``ladder'') were ranked as the easiest.

\begin{figure}[hbt!]
    \centering
    \includegraphics[width=0.95\linewidth]{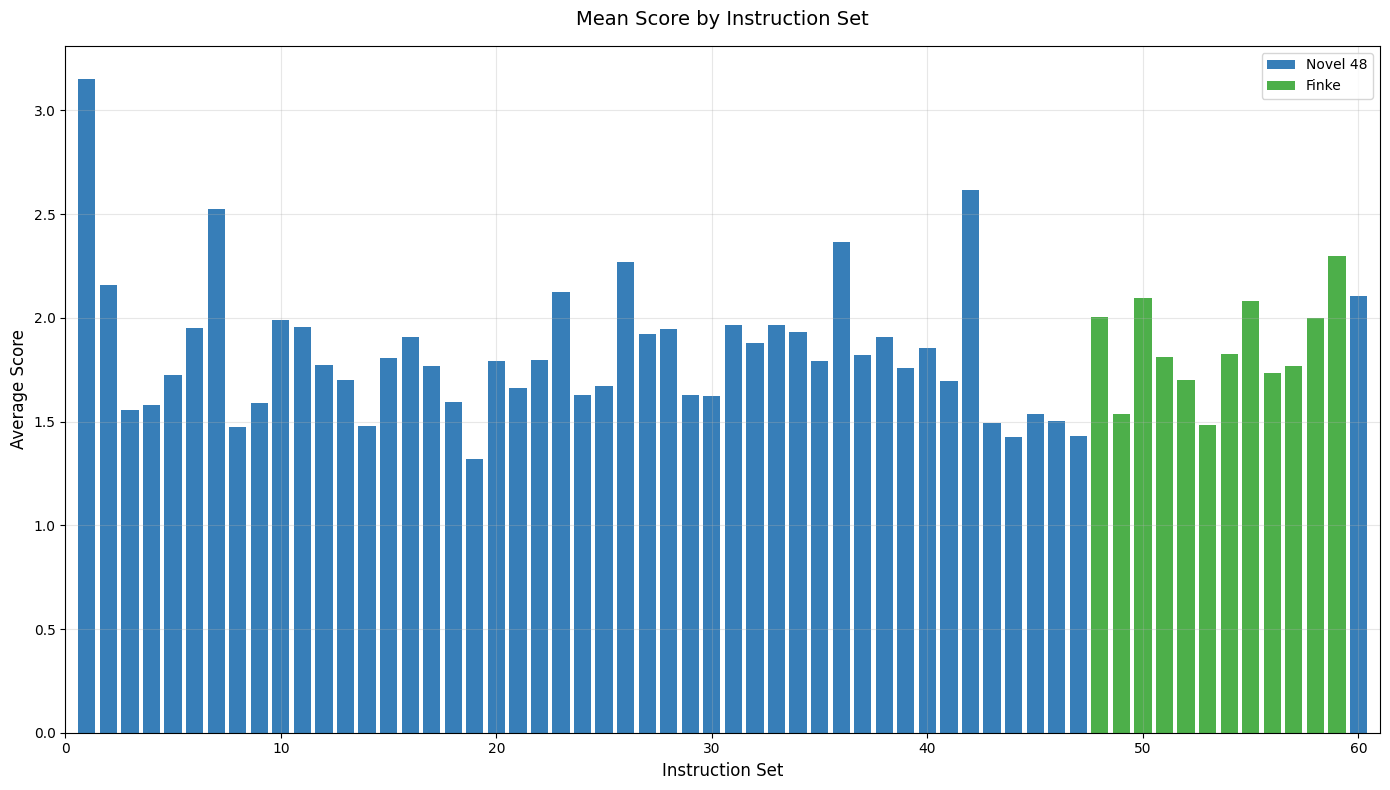}
    \caption{The mean crowd-sourced score for each answer given in each instruction set. Particularly difficult trials had a large variety of answers, many with lower scores.}
    \label{fig:mean_score_instruction_set}
\end{figure}

We can validate this by viewing the mean score given to each instruction set by the graders, and seeing a broad distribution (Supplemental Figure \ref{fig:mean_score_instruction_set}).

\subsection{VVIQ Analysis} \label{sec:vviq}

The VVIQ scale range is 16 (minimum) to 80 (maximum). Subjects had a mean VVIQ score of 55.8, confirming that our sample was normal given recent large sampling efforts \citep{wright2024}. We did not find any correlation between VVIQ scores and subjects' performance (Pearson's $r=-0.17$, $p=.0942$). If anything, there was a negative trend, suggesting that imagery capacity as measured by the VVIQ could not predict performance in a task designed to measure mental imagery. See Supplemental Figure \ref{fig:vviq_corr}. 

Of the 100 subjects we tested, one subject qualified as a true aphantasic (VVIQ score of 16, the minimum). This subject, however, had the 5th highest score of all human subjects. While this constitutes a single data point (all other subjects had higher VVIQ scores), it raises the question of how this subject was able to complete the task. Recent data has shown that aphantasics can complete imagery tasks surprisingly well \citep{blomkvist2023, kay2024, pounder2022}, even if the exact way in which they accomplish this or whether they use a single strategy across tasks is still unclear. 

There were three subjects who qualified as low-imagers (or weak aphantasics) scoring between 17 and 32 on the VVIQ (23, 29, 29). Two of the subjects performed around the median score for humans (53rd and 56th), while the third was the 15th highest scoring individual in our dataset.

Lastly, we had seven subjects who had very high imagery (or hyperphantasics) scoring between 75 and 80 (the highest) on the VVIQ (one scored 77 the rest 80). These subjects had very mixed performance with three subjects performing above the median (18th, 20th, and 24th), and four subjects performing well below the median, and in a couple cases getting close to the bottom performance (77th, 81st, 91st, and 94th).

The high performance exhibited by the aphantasic subjects and the most advanced LLMs, in addition to the lack of correlation between imagery capacity (measured via VVIQ) and performance in the task, offers further evidence that mental imagery may operate in a propositional format. At the very least, this supposedly gold standard for probing pictorial imagining may not be as well suited for the task as generations of cognitive psychologists have thought. Naturally, much more data and analysis is needed to cement this conclusion. The current experiments with LLMs, and future ones as well, may help further our understanding of how aphantasics are able to accomplish these and other tasks. (For one possible explanation appealing to spatial imagery, see section \ref{sec:discussion} in the main text).

\section{Technical Setup} \label{sec:technicalsetup}

All model prompting was done through the respective APIs of the models' parent company in Python. Models were run at least once in multiple-context with most models being run at least once in both single-context and multiple-context. After determining that there was no statistical difference between the two paradigms, any subsequent runs of models were run in multiple-context (due to the decreased cost of inference because of less input token usage).

Our exact runs of model paradigms are as follows:
\begin{itemize}
    \item Claude Opus 4.1, multiple-context: 1
    \item Claude Sonnet 4, multiple-context: 1
    \item Claude Sonnet 4, single-context: 1
    \item Gemini 2.0 Flash, multiple-context: 1
    \item Gemini 2.0 Flash, single-context: 1
    \item Gemini 2.0 Flash images, multiple-context: 1
    \item Gemini 2.5 Pro, multiple-context: 1
    \item Gemini 2.5 Pro, single-context: 1
    \item ChatGPT-4o, multiple-context: 1
    \item ChatGPT-4o, single-context: 1
    \item GPT-4.1, multiple-context: 1
    \item GPT-4.1, single-context: 1
    \item GPT-4.1 with GPT-image-1, multiple-context: 1
    \item GPT-4.1 with GPT-image-1, single-context: 1
    \item GPT-5, high reasoning, multiple-context: 2
    \item GPT-5, medium reasoning, multiple-context: 1
    \item GPT-5, low reasoning, multiple-context: 1
    \item GPT-5, minimal reasoning, multiple-context: 1
    \item o3, high reasoning, multiple-context: 2
    \item o3, high reasoning, single-context: 1
    \item o3, medium reasoning, multiple-context: 1
    \item o3, low reasoning, multiple-context: 1
    \item o3-Pro, high reasoning, multiple-context: 2
    \item o3-Pro, high reasoning, single-context: 1
    \item o3 with GPT-image-1, high reasoning, multiple-context: 4
    \item o3 with GPT-image-1, medium reasoning, multiple-context: 1
    \item o4-mini, high reasoning, multiple-context: 1
    \item o4-mini, high reasoning, single-context: 1
    \item o4-mini, medium reasoning, multiple-context: 1
    \item o4-mini, medium reasoning, single-context: 1
\end{itemize}

The strongest OpenAI reasoning models were run a single additional time with variations on the `reasoning\_effort' parameter as described in Section \ref{sec:reasoning}. Our reported data in all other sections is based on the `reasoning\_effort' parameter being set to `high'.

For all models which allowed modification in their `temperature' parameter, we set it to 0.1 (reasoning models from OpenAI and Anthropic do not allow modification to temperature). All other hyperparameters were left at their default value.

For OpenAI, we used the openai package, version 1.76.2; for Gemini, the google-genai package version 1.14.0; for Claude, the anthropic package version 0.52.2. We used the Miniconda environment manager to create a Python 3.12.9 environment, and the complete frozen package list (`python\_env.yml') is available in our repository (see Appendix \ref{sec:code_and_data}).

Our prompt processing pipeline was run on a 2019 Razer Blade Stealth running Arch Linux with 16 GB of RAM, an 8 core Intel CPU (i7-8565U), and no dedicated GPU.

All data collected from models totaled around \$1000 US in cost. The exact versions of models is listed in Table \ref{tab:modelversions}.

Data pre-processing was performed in a Jupyter Notebook file. We provide .md and .html versions of the output in addition to the raw file for convenience.

Statistical analyses were performed using R 4.3.3 in PyCharm with the R plugin installed. Miniconda was also used to create a replicable environment and the frozen package list (`r\_env.yml') is available in our repository. Our analyses were performed within an R Markdown script. We provide .html and .pdf exports for simplified viewing.






\clearpage
\section{Supplemental Figures}

\begin{figure}[hbt!]
\centering
\includegraphics[width=0.95\linewidth]{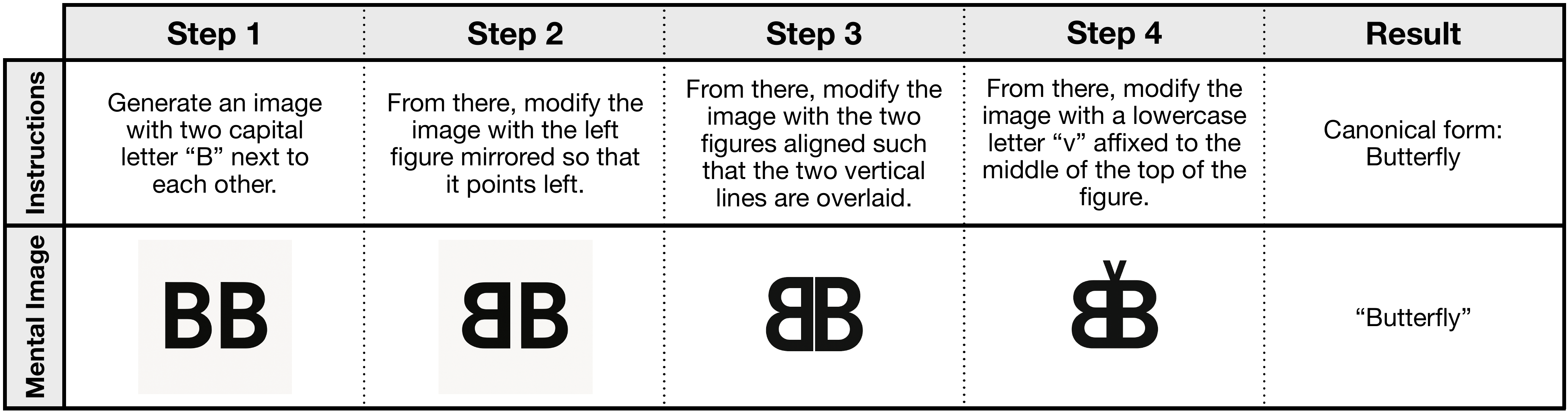}
\caption{An example image generation output from GPT-image-1 in combination with o3. No seed image was given to the model, but subsequent steps retained the previous image and asked for its modification.}
\label{fig:imageoutput}
\end{figure}

\begin{figure}[hbt!]
    \centering
    \includegraphics[width=0.95\linewidth]{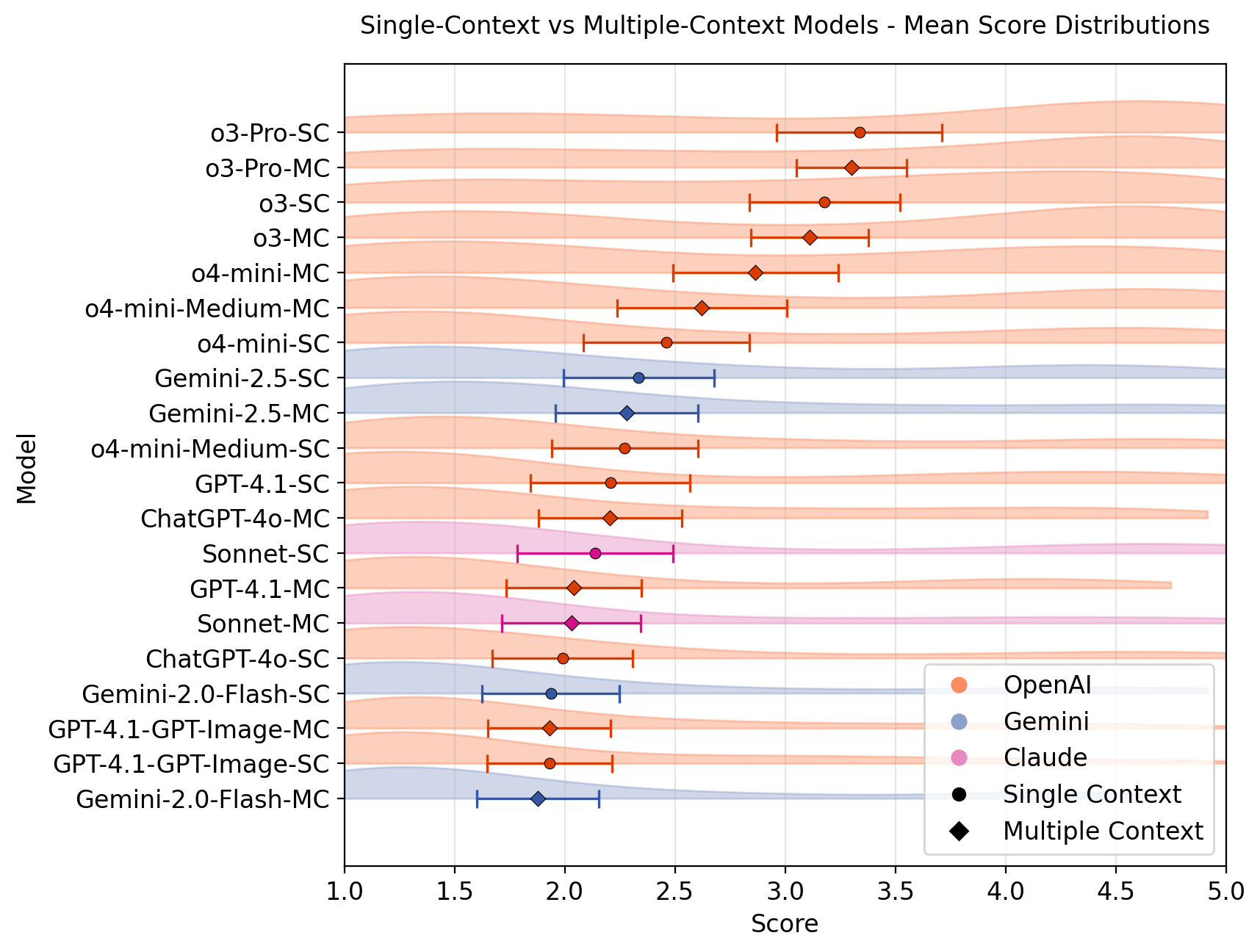}
    \caption{95\% confidence intervals showing the differences between Single-Context and Multiple-Context. Single-Context has a slight, non-significant edge in most cases. All context variant comparisons were non-significant after correcting for multiple-comparisons.}
    \label{fig:sc_mc_ci}
\end{figure}



\begin{figure}[hbt!]
    \centering
    \includegraphics[width=0.95\linewidth]{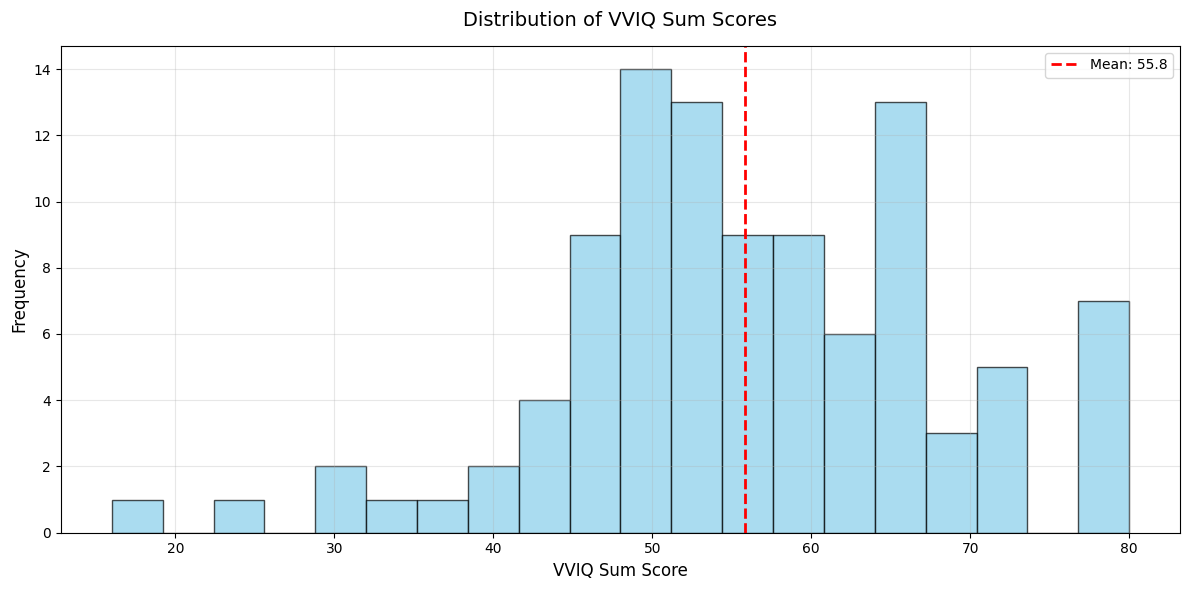}
    \caption{The distribution of VVIQ scores for our 100 human subjects. 1 subject qualified as an aphantasic under the strictest condition (VVIQ = 16). The characteristic left skew of the distribution is clear.}
    \label{fig:vviq_dist}
\end{figure}




\begin{figure}[hbt!]
    \centering
    \includegraphics[width=0.95\linewidth]{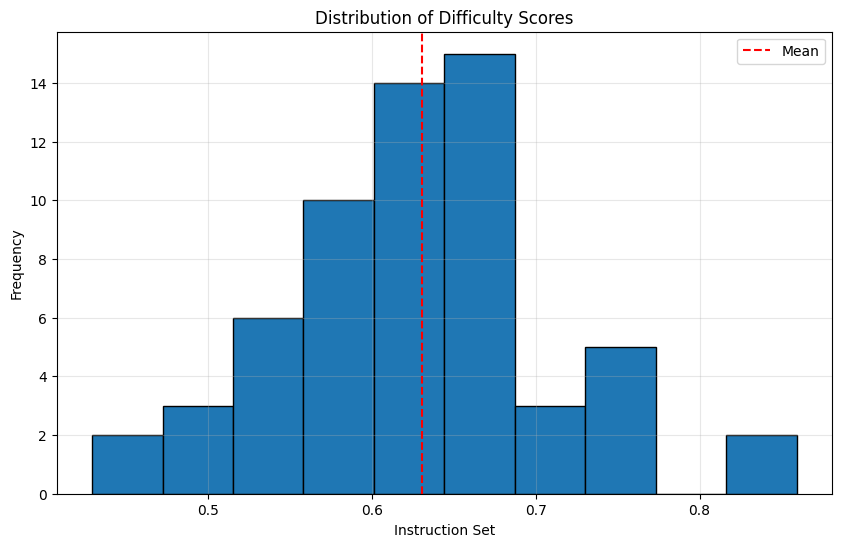}
    \caption{Distribution of calculated difficulty scores per instruction set. Our instruction sets showed strong variance of difficulty.}
    \label{fig:diff_dist}
\end{figure}

\begin{figure}[hbt!]
    \centering
    \includegraphics[width=0.95\linewidth]{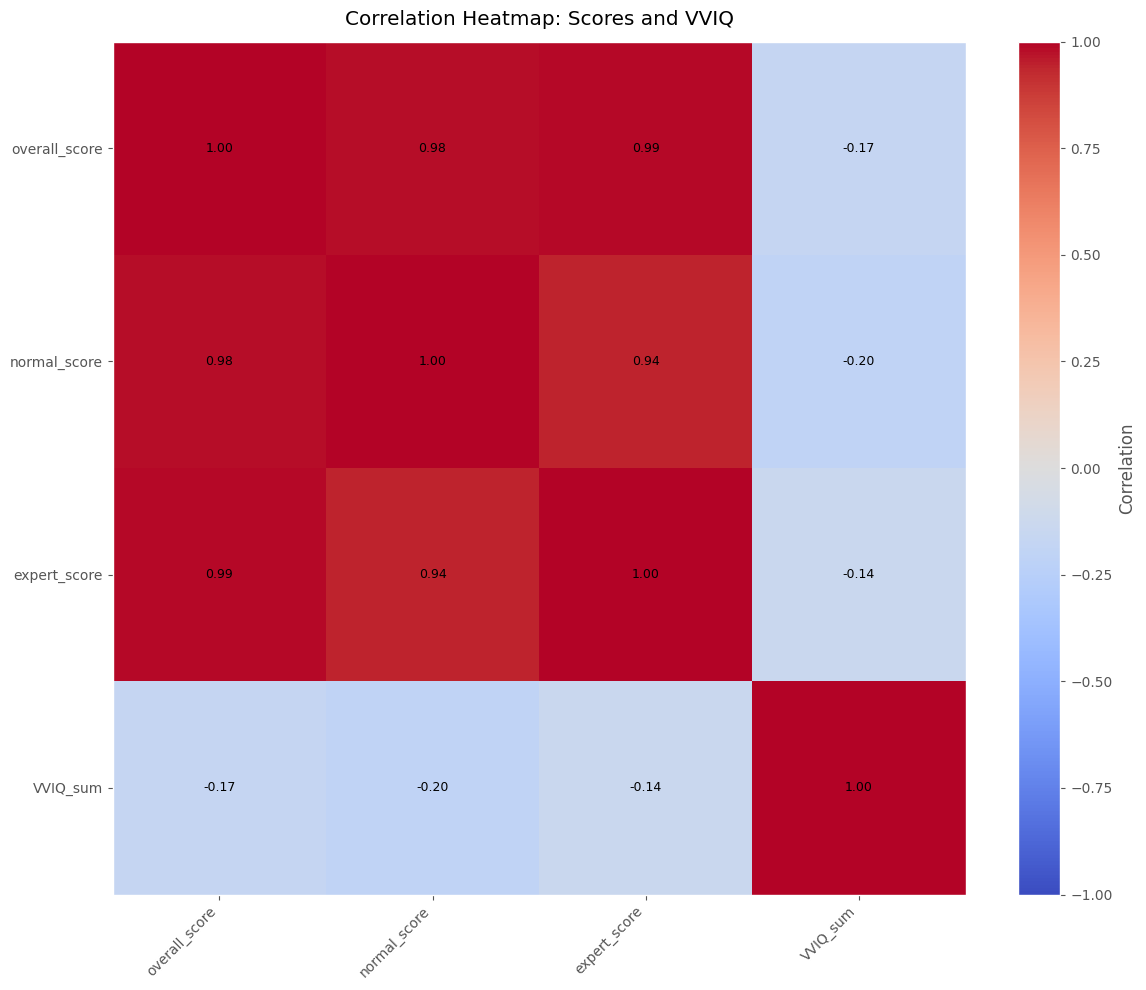}
    \caption{Correlation between VVIQ score sum, `overall\_score', `normal\_score', and `expert\_score' (see Appendix \ref{sec:grade_weight} for terminology explanation).}
    \label{fig:vviq_corr}
\end{figure}

\clearpage
\section{Supplemental Tables}

\begin{table}[hbt]
\centering
\begingroup
\caption{Model selection and features. For OpenAI reasoning models, `reasoning\_effort' was set to `high' (as shown) and, for the models indicated, GPT-image-1 was integrated. For Gemini models, default parameters regarding reasoning were retained and native image generation tools were used. For Claude models a reasoning token budget was manually given (well above what was ever allocated). This table only shows the primarily graded models, not the reasoning variations in our reasoning model effort analysis.}
\vspace{1em} 
\label{tab:modelselection}
\begin{tabular}{ |l|c|c| }
\hline
\rowcolor{blue!25} Model             & Reasoning & Image Generation \\ 
\hline
GPT-5             & High     & No \\
\hline
o3-Pro            & High     & No \\
\hline
o3                & High     & Yes \\
\hline
o4-mini           & High     & No \\
\hline
ChatGPT-4o        & No      & No \\
\hline
GPT-4.1           & No      & Yes \\
\hline
Gemini 2.5 Pro    & Dynamic & No \\ 
\hline
Gemini 2.0 Flash  & No      & Yes \\ 
\hline
Claude Sonnet 4   & 4000t   & No \\
\hline
Claude Opus 4.1   & 9000t   & No \\
\hline
Gemini 3 Pro      & High    & No \\
\hline
\end{tabular}
\endgroup
\end{table}

\begin{table}[hbt]
\centering
\caption{Model Versions. Model versions used in our analysis. When possible exact dated versions are given, if no such version is available the date of usage was also provided.}
\begingroup
\renewcommand{\arraystretch}{1.25} 

\begin{tabular}{ |l|l| } 
\hline
\rowcolor{blue!25} Model & Version \\ 
GPT-5             & gpt-5-2025-08-07 \\
\hline
o3                & o3-2025-04-16 \\
\hline
o3-Pro            & o3-pro-2025-06-10    \\
\hline
o4-mini           & o4-mini-2025-04-16    \\
\hline
ChatGPT-4o        & chatgpt-4o-latest (July 2025)  \\
\hline
GPT-4.1           & gpt-4\_1-2025-04-14      \\
\hline
GPT-image-1       & gpt-image-1-2025-04-23      \\
\hline
Gemini 2.0 Flash  & gemini-2.0-flash (February 2025) \\
\hline
2.0 Flash Image Preview    & gemini-2.0-flash-preview-image-generation (May 2025) \\
\hline
Gemini 2.5 Pro    & gemini-2.5-pro-preview-05-06 \\ 
\hline
Claude Sonnet 4   & claude-sonnet-4-20250514 \\
\hline
Claude Opus 4.1   & claude-opus-4-1-20250805 \\
\hline
Gemini 3 Pro   & gemini-3-pro-preview (November 2025) \\
\end{tabular}
\endgroup
\label{tab:modelversions}
\end{table} 

\begin{table}[hbt]
\centering
\begingroup
\caption{Reasoning Variations. Reasoning level variations used for reasoning analysis.}
\vspace{1em} 
\label{tab:reasoning_variations}
\begin{tabular}{ |l|c|c| }
\hline
\rowcolor{blue!25} Model  & Reasoning Levels Used \\ 
\hline
GPT-5             & Minimal, Low, Medium, High\\
\hline
o3-Pro            & High    \\
\hline
o3                & Low, Medium, High     \\
\hline
o3 with GPT-image-1 & Medium, High \\
\hline
o4-mini           & Medium, High \\
\hline
\end{tabular}
\endgroup
\end{table}
\clearpage

\end{appendices}



\bibliography{iclr2026_conference}

@article{kaski2002,
author = {Diego Kaski},
title ={Revision: Is Visual Perception a Requisite for Visual Imagery?},

journal = {Perception},
volume = {31},
number = {6},
pages = {717-731},
year = {2002},
doi = {10.1068/p3360},
    note ={PMID: 12092798},

URL = { 
    
        https://doi.org/10.1068/p3360
    
    

},
eprint = { 
    
        https://doi.org/10.1068/p3360
    
    

}
}

@InProceedings{ma2025,
    author    = {Ma, Wufei and Chen, Haoyu and Zhang, Guofeng and Chou, Yu-Cheng and Chen, Jieneng and de Melo, Celso and Yuille, Alan},
    title     = {3DSRBench: A Comprehensive 3D Spatial Reasoning Benchmark},
    booktitle = {Proceedings of the IEEE/CVF International Conference on Computer Vision (ICCV)},
    month     = {October},
    year      = {2025},
    pages     = {6924-6934}
}

@misc{chollet2025,
      title={ARC Prize 2024: Technical Report}, 
      author={Francois Chollet and Mike Knoop and Gregory Kamradt and Bryan Landers},
      year={2025},
      eprint={2412.04604},
      archivePrefix={arXiv},
      primaryClass={cs.AI},
      url={https://arxiv.org/abs/2412.04604}, 
      accessed={11 March 2026}
}

@misc{vatsa2025,
      title={Right Looks, Wrong Reasons: Compositional Fidelity in Text-to-Image Generation}, 
      author={Mayank Vatsa and Aparna Bharati and Richa Singh},
      year={2025},
      eprint={2511.10136},
      archivePrefix={arXiv},
      primaryClass={cs.CV},
      url={https://arxiv.org/abs/2511.10136}, 
      accessed={9 March 2026}
}

@article{
pearson2015,
author = {Joel Pearson  and Stephen M. Kosslyn },
title = {The heterogeneity of mental representation: Ending the imagery debate},
journal = {Proceedings of the National Academy of Sciences},
volume = {112},
number = {33},
pages = {10089-10092},
year = {2015},
doi = {10.1073/pnas.1504933112},
URL = {https://www.pnas.org/doi/abs/10.1073/pnas.1504933112},
eprint = {https://www.pnas.org/doi/pdf/10.1073/pnas.1504933112}}

@inproceedings{ismayilzada2025,
    title = "Evaluating Morphological Compositional Generalization in Large Language Models",
    author = {Ismayilzada, Mete  and
      Circi, Defne  and
      S{\"a}lev{\"a}, Jonne  and
      Sirin, Hale  and
      K{\"o}ksal, Abdullatif  and
      Dhingra, Bhuwan  and
      Bosselut, Antoine  and
      Ataman, Duygu  and
      Plas, Lonneke Van Der},
    editor = "Chiruzzo, Luis  and
      Ritter, Alan  and
      Wang, Lu",
    booktitle = "Proceedings of the 2025 Conference of the Nations of the Americas Chapter of the Association for Computational Linguistics: Human Language Technologies (Volume 1: Long Papers)",
    month = apr,
    year = "2025",
    address = "Albuquerque, New Mexico",
    publisher = "Association for Computational Linguistics",
    url = "https://aclanthology.org/2025.naacl-long.59/",
    doi = "10.18653/v1/2025.naacl-long.59",
    pages = "1270--1305",
    ISBN = "979-8-89176-189-6"
}

@misc{khandelwal2025,
      title={How Do Language Models Compose Functions?}, 
      author={Apoorv Khandelwal and Ellie Pavlick},
      year={2025},
      eprint={2510.01685},
      archivePrefix={arXiv},
      primaryClass={cs.CL},
      url={https://arxiv.org/abs/2510.01685}, 
      accessed={9 March 2026}
}

@book{fodor1975,
  title={The language of thought},
  author={Fodor, Jerry A},
  volume={5},
  year={1975},
  publisher={Harvard university press}
}

@inproceedings{sun2025,
    title = "Are We in the {AI}-Generated Text World Already? Quantifying and Monitoring {AIGT} on Social Media",
    author = "Sun, Zhen  and
      Zhang, Zongmin  and
      Shen, Xinyue  and
      Zhang, Ziyi  and
      Liu, Yule  and
      Backes, Michael  and
      Zhang, Yang  and
      He, Xinlei",
    editor = "Che, Wanxiang  and
      Nabende, Joyce  and
      Shutova, Ekaterina  and
      Pilehvar, Mohammad Taher",
    booktitle = "Proceedings of the 63rd Annual Meeting of the Association for Computational Linguistics (Volume 1: Long Papers)",
    month = jul,
    year = "2025",
    address = "Vienna, Austria",
    publisher = "Association for Computational Linguistics",
    url = "https://aclanthology.org/2025.acl-long.1120/",
    doi = "10.18653/v1/2025.acl-long.1120",
    pages = "22975--23005",
    ISBN = "979-8-89176-251-0"
}

@misc{ghosh2025,
      title={Exploring the Frontier of Vision-Language Models: A Survey of Current Methodologies and Future Directions}, 
      author={Akash Ghosh and Arkadeep Acharya and Sriparna Saha and Vinija Jain and Aman Chadha},
      year={2025},
      eprint={2404.07214},
      archivePrefix={arXiv},
      primaryClass={cs.CV},
      url={https://arxiv.org/abs/2404.07214}, 
      accessed={9 March 2026}
}

@misc{qwen3tech,
      title={Qwen3 Technical Report}, 
      author={An Yang and Anfeng Li and Baosong Yang and Beichen Zhang and Binyuan Hui and Bo Zheng and Bowen Yu and Chang Gao and Chengen Huang and Chenxu Lv and Chujie Zheng and Dayiheng Liu and Fan Zhou and Fei Huang and Feng Hu and Hao Ge and Haoran Wei and Huan Lin and Jialong Tang and Jian Yang and Jianhong Tu and Jianwei Zhang and Jianxin Yang and Jiaxi Yang and Jing Zhou and Jingren Zhou and Junyang Lin and Kai Dang and Keqin Bao and Kexin Yang and Le Yu and Lianghao Deng and Mei Li and Mingfeng Xue and Mingze Li and Pei Zhang and Peng Wang and Qin Zhu and Rui Men and Ruize Gao and Shixuan Liu and Shuang Luo and Tianhao Li and Tianyi Tang and Wenbiao Yin and Xingzhang Ren and Xinyu Wang and Xinyu Zhang and Xuancheng Ren and Yang Fan and Yang Su and Yichang Zhang and Yinger Zhang and Yu Wan and Yuqiong Liu and Zekun Wang and Zeyu Cui and Zhenru Zhang and Zhipeng Zhou and Zihan Qiu},
      year={2025},
      eprint={2505.09388},
      archivePrefix={arXiv},
      primaryClass={cs.CL},
      url={https://arxiv.org/abs/2505.09388}, 
      accessed={9 March 2026}
}

@misc{qwen25tech,
      title={Qwen2.5 Technical Report}, 
      author={Qwen and : and An Yang and Baosong Yang and Beichen Zhang and Binyuan Hui and Bo Zheng and Bowen Yu and Chengyuan Li and Dayiheng Liu and Fei Huang and Haoran Wei and Huan Lin and Jian Yang and Jianhong Tu and Jianwei Zhang and Jianxin Yang and Jiaxi Yang and Jingren Zhou and Junyang Lin and Kai Dang and Keming Lu and Keqin Bao and Kexin Yang and Le Yu and Mei Li and Mingfeng Xue and Pei Zhang and Qin Zhu and Rui Men and Runji Lin and Tianhao Li and Tianyi Tang and Tingyu Xia and Xingzhang Ren and Xuancheng Ren and Yang Fan and Yang Su and Yichang Zhang and Yu Wan and Yuqiong Liu and Zeyu Cui and Zhenru Zhang and Zihan Qiu},
      year={2025},
      eprint={2412.15115},
      archivePrefix={arXiv},
      primaryClass={cs.CL},
      url={https://arxiv.org/abs/2412.15115}, 
      accessed={9 March 2026}
}

@misc{gpt5tech,
      title={OpenAI GPT-5 System Card}, 
      author={Aaditya Singh and Adam Fry and Adam Perelman and Adam Tart and Adi Ganesh and Ahmed El-Kishky and Aidan McLaughlin and Aiden Low and AJ Ostrow and Akhila Ananthram and Akshay Nathan and Alan Luo and Alec Helyar and Aleksander Madry and Aleksandr Efremov and Aleksandra Spyra and Alex Baker-Whitcomb and Alex Beutel and Alex Karpenko and Alex Makelov and Alex Neitz and Alex Wei and Alexandra Barr and Alexandre Kirchmeyer and Alexey Ivanov and Alexi Christakis and Alistair Gillespie and Allison Tam and Ally Bennett and Alvin Wan and Alyssa Huang and Amy McDonald Sandjideh and Amy Yang and Ananya Kumar and Andre Saraiva and Andrea Vallone and Andrei Gheorghe and Andres Garcia Garcia and Andrew Braunstein and Andrew Liu and Andrew Schmidt and Andrey Mereskin and Andrey Mishchenko and Andy Applebaum and Andy Rogerson and Ann Rajan and Annie Wei and Anoop Kotha and Anubha Srivastava and Anushree Agrawal and Arun Vijayvergiya and Ashley Tyra and Ashvin Nair and Avi Nayak and Ben Eggers and Bessie Ji and Beth Hoover and Bill Chen and Blair Chen and Boaz Barak and Borys Minaiev and Botao Hao and Bowen Baker and Brad Lightcap and Brandon McKinzie and Brandon Wang and Brendan Quinn and Brian Fioca and Brian Hsu and Brian Yang and Brian Yu and Brian Zhang and Brittany Brenner and Callie Riggins Zetino and Cameron Raymond and Camillo Lugaresi and Carolina Paz and Cary Hudson and Cedric Whitney and Chak Li and Charles Chen and Charlotte Cole and Chelsea Voss and Chen Ding and Chen Shen and Chengdu Huang and Chris Colby and Chris Hallacy and Chris Koch and Chris Lu and Christina Kaplan and Christina Kim and CJ Minott-Henriques and Cliff Frey and Cody Yu and Coley Czarnecki and Colin Reid and Colin Wei and Cory Decareaux and Cristina Scheau and Cyril Zhang and Cyrus Forbes and Da Tang and Dakota Goldberg and Dan Roberts and Dana Palmie and Daniel Kappler and Daniel Levine and Daniel Wright and Dave Leo and David Lin and David Robinson and Declan Grabb and Derek Chen and Derek Lim and Derek Salama and Dibya Bhattacharjee and Dimitris Tsipras and Dinghua Li and Dingli Yu and DJ Strouse and Drew Williams and Dylan Hunn and Ed Bayes and Edwin Arbus and Ekin Akyurek and Elaine Ya Le and Elana Widmann and Eli Yani and Elizabeth Proehl and Enis Sert and Enoch Cheung and Eri Schwartz and Eric Han and Eric Jiang and Eric Mitchell and Eric Sigler and Eric Wallace and Erik Ritter and Erin Kavanaugh and Evan Mays and Evgenii Nikishin and Fangyuan Li and Felipe Petroski Such and Filipe de Avila Belbute Peres and Filippo Raso and Florent Bekerman and Foivos Tsimpourlas and Fotis Chantzis and Francis Song and Francis Zhang and Gaby Raila and Garrett McGrath and Gary Briggs and Gary Yang and Giambattista Parascandolo and Gildas Chabot and Grace Kim and Grace Zhao and Gregory Valiant and Guillaume Leclerc and Hadi Salman and Hanson Wang and Hao Sheng and Haoming Jiang and Haoyu Wang and Haozhun Jin and Harshit Sikchi and Heather Schmidt and Henry Aspegren and Honglin Chen and Huida Qiu and Hunter Lightman and Ian Covert and Ian Kivlichan and Ian Silber and Ian Sohl and Ibrahim Hammoud and Ignasi Clavera and Ikai Lan and Ilge Akkaya and Ilya Kostrikov and Irina Kofman and Isak Etinger and Ishaan Singal and Jackie Hehir and Jacob Huh and Jacqueline Pan and Jake Wilczynski and Jakub Pachocki and James Lee and James Quinn and Jamie Kiros and Janvi Kalra and Jasmyn Samaroo and Jason Wang and Jason Wolfe and Jay Chen and Jay Wang and Jean Harb and Jeffrey Han and Jeffrey Wang and Jennifer Zhao and Jeremy Chen and Jerene Yang and Jerry Tworek and Jesse Chand and Jessica Landon and Jessica Liang and Ji Lin and Jiancheng Liu and Jianfeng Wang and Jie Tang and Jihan Yin and Joanne Jang and Joel Morris and Joey Flynn and Johannes Ferstad and Johannes Heidecke and John Fishbein and John Hallman and Jonah Grant and Jonathan Chien and Jonathan Gordon and Jongsoo Park and Jordan Liss and Jos Kraaijeveld and Joseph Guay and Joseph Mo and Josh Lawson and Josh McGrath and Joshua Vendrow and Joy Jiao and Julian Lee and Julie Steele and Julie Wang and Junhua Mao and Kai Chen and Kai Hayashi and Kai Xiao and Kamyar Salahi and Kan Wu and Karan Sekhri and Karan Sharma and Karan Singhal and Karen Li and Kenny Nguyen and Keren Gu-Lemberg and Kevin King and Kevin Liu and Kevin Stone and Kevin Yu and Kristen Ying and Kristian Georgiev and Kristie Lim and Kushal Tirumala and Kyle Miller and Lama Ahmad and Larry Lv and Laura Clare and Laurance Fauconnet and Lauren Itow and Lauren Yang and Laurentia Romaniuk and Leah Anise and Lee Byron and Leher Pathak and Leon Maksin and Leyan Lo and Leyton Ho and Li Jing and Liang Wu and Liang Xiong and Lien Mamitsuka and Lin Yang and Lindsay McCallum and Lindsey Held and Liz Bourgeois and Logan Engstrom and Lorenz Kuhn and Louis Feuvrier and Lu Zhang and Lucas Switzer and Lukas Kondraciuk and Lukasz Kaiser and Manas Joglekar and Mandeep Singh and Mandip Shah and Manuka Stratta and Marcus Williams and Mark Chen and Mark Sun and Marselus Cayton and Martin Li and Marvin Zhang and Marwan Aljubeh and Matt Nichols and Matthew Haines and Max Schwarzer and Mayank Gupta and Meghan Shah and Melody Huang and Meng Dong and Mengqing Wang and Mia Glaese and Micah Carroll and Michael Lampe and Michael Malek and Michael Sharman and Michael Zhang and Michele Wang and Michelle Pokrass and Mihai Florian and Mikhail Pavlov and Miles Wang and Ming Chen and Mingxuan Wang and Minnia Feng and Mo Bavarian and Molly Lin and Moose Abdool and Mostafa Rohaninejad and Nacho Soto and Natalie Staudacher and Natan LaFontaine and Nathan Marwell and Nelson Liu and Nick Preston and Nick Turley and Nicklas Ansman and Nicole Blades and Nikil Pancha and Nikita Mikhaylin and Niko Felix and Nikunj Handa and Nishant Rai and Nitish Keskar and Noam Brown and Ofir Nachum and Oleg Boiko and Oleg Murk and Olivia Watkins and Oona Gleeson and Pamela Mishkin and Patryk Lesiewicz and Paul Baltescu and Pavel Belov and Peter Zhokhov and Philip Pronin and Phillip Guo and Phoebe Thacker and Qi Liu and Qiming Yuan and Qinghua Liu and Rachel Dias and Rachel Puckett and Rahul Arora and Ravi Teja Mullapudi and Raz Gaon and Reah Miyara and Rennie Song and Rishabh Aggarwal and RJ Marsan and Robel Yemiru and Robert Xiong and Rohan Kshirsagar and Rohan Nuttall and Roman Tsiupa and Ronen Eldan and Rose Wang and Roshan James and Roy Ziv and Rui Shu and Ruslan Nigmatullin and Saachi Jain and Saam Talaie and Sam Altman and Sam Arnesen and Sam Toizer and Sam Toyer and Samuel Miserendino and Sandhini Agarwal and Sarah Yoo and Savannah Heon and Scott Ethersmith and Sean Grove and Sean Taylor and Sebastien Bubeck and Sever Banesiu and Shaokyi Amdo and Shengjia Zhao and Sherwin Wu and Shibani Santurkar and Shiyu Zhao and Shraman Ray Chaudhuri and Shreyas Krishnaswamy and Shuaiqi and Xia and Shuyang Cheng and Shyamal Anadkat and Simón Posada Fishman and Simon Tobin and Siyuan Fu and Somay Jain and Song Mei and Sonya Egoian and Spencer Kim and Spug Golden and SQ Mah and Steph Lin and Stephen Imm and Steve Sharpe and Steve Yadlowsky and Sulman Choudhry and Sungwon Eum and Suvansh Sanjeev and Tabarak Khan and Tal Stramer and Tao Wang and Tao Xin and Tarun Gogineni and Taya Christianson and Ted Sanders and Tejal Patwardhan and Thomas Degry and Thomas Shadwell and Tianfu Fu and Tianshi Gao and Timur Garipov and Tina Sriskandarajah and Toki Sherbakov and Tomer Kaftan and Tomo Hiratsuka and Tongzhou Wang and Tony Song and Tony Zhao and Troy Peterson and Val Kharitonov and Victoria Chernova and Vineet Kosaraju and Vishal Kuo and Vitchyr Pong and Vivek Verma and Vlad Petrov and Wanning Jiang and Weixing Zhang and Wenda Zhou and Wenlei Xie and Wenting Zhan and Wes McCabe and Will DePue and Will Ellsworth and Wulfie Bain and Wyatt Thompson and Xiangning Chen and Xiangyu Qi and Xin Xiang and Xinwei Shi and Yann Dubois and Yaodong Yu and Yara Khakbaz and Yifan Wu and Yilei Qian and Yin Tat Lee and Yinbo Chen and Yizhen Zhang and Yizhong Xiong and Yonglong Tian and Young Cha and Yu Bai and Yu Yang and Yuan Yuan and Yuanzhi Li and Yufeng Zhang and Yuguang Yang and Yujia Jin and Yun Jiang and Yunyun Wang and Yushi Wang and Yutian Liu and Zach Stubenvoll and Zehao Dou and Zheng Wu and Zhigang Wang},
      year={2025},
      eprint={2601.03267},
      archivePrefix={arXiv},
      primaryClass={cs.CL},
      url={https://arxiv.org/abs/2601.03267}, 
      accessed={9 March 2026}
}

@misc{geminitech,
      title={Gemini: A Family of Highly Capable Multimodal Models}, 
      author={Gemini Team and Rohan Anil and Sebastian Borgeaud and Jean-Baptiste Alayrac and Jiahui Yu and Radu Soricut and Johan Schalkwyk and Andrew M. Dai and Anja Hauth and Katie Millican and David Silver and Melvin Johnson and Ioannis Antonoglou and Julian Schrittwieser and Amelia Glaese and Jilin Chen and Emily Pitler and Timothy Lillicrap and Angeliki Lazaridou and Orhan Firat and James Molloy and Michael Isard and Paul R. Barham and Tom Hennigan and Benjamin Lee and Fabio Viola and Malcolm Reynolds and Yuanzhong Xu and Ryan Doherty and Eli Collins and Clemens Meyer and Eliza Rutherford and Erica Moreira and Kareem Ayoub and Megha Goel and Jack Krawczyk and Cosmo Du and Ed Chi and Heng-Tze Cheng and Eric Ni and Purvi Shah and Patrick Kane and Betty Chan and Manaal Faruqui and Aliaksei Severyn and Hanzhao Lin and YaGuang Li and Yong Cheng and Abe Ittycheriah and Mahdis Mahdieh and Mia Chen and Pei Sun and Dustin Tran and Sumit Bagri and Balaji Lakshminarayanan and Jeremiah Liu and Andras Orban and Fabian Güra and Hao Zhou and Xinying Song and Aurelien Boffy and Harish Ganapathy and Steven Zheng and HyunJeong Choe and Ágoston Weisz and Tao Zhu and Yifeng Lu and Siddharth Gopal and Jarrod Kahn and Maciej Kula and Jeff Pitman and Rushin Shah and Emanuel Taropa and Majd Al Merey and Martin Baeuml and Zhifeng Chen and Laurent El Shafey and Yujing Zhang and Olcan Sercinoglu and George Tucker and Enrique Piqueras and Maxim Krikun and Iain Barr and Nikolay Savinov and Ivo Danihelka and Becca Roelofs and Anaïs White and Anders Andreassen and Tamara von Glehn and Lakshman Yagati and Mehran Kazemi and Lucas Gonzalez and Misha Khalman and Jakub Sygnowski and Alexandre Frechette and Charlotte Smith and Laura Culp and Lev Proleev and Yi Luan and Xi Chen and James Lottes and Nathan Schucher and Federico Lebron and Alban Rrustemi and Natalie Clay and Phil Crone and Tomas Kocisky and Jeffrey Zhao and Bartek Perz and Dian Yu and Heidi Howard and Adam Bloniarz and Jack W. Rae and Han Lu and Laurent Sifre and Marcello Maggioni and Fred Alcober and Dan Garrette and Megan Barnes and Shantanu Thakoor and Jacob Austin and Gabriel Barth-Maron and William Wong and Rishabh Joshi and Rahma Chaabouni and Deeni Fatiha and Arun Ahuja and Gaurav Singh Tomar and Evan Senter and Martin Chadwick and Ilya Kornakov and Nithya Attaluri and Iñaki Iturrate and Ruibo Liu and Yunxuan Li and Sarah Cogan and Jeremy Chen and Chao Jia and Chenjie Gu and Qiao Zhang and Jordan Grimstad and Ale Jakse Hartman and Xavier Garcia and Thanumalayan Sankaranarayana Pillai and Jacob Devlin and Michael Laskin and Diego de Las Casas and Dasha Valter and Connie Tao and Lorenzo Blanco and Adrià Puigdomènech Badia and David Reitter and Mianna Chen and Jenny Brennan and Clara Rivera and Sergey Brin and Shariq Iqbal and Gabriela Surita and Jane Labanowski and Abhi Rao and Stephanie Winkler and Emilio Parisotto and Yiming Gu and Kate Olszewska and Ravi Addanki and Antoine Miech and Annie Louis and Denis Teplyashin and Geoff Brown and Elliot Catt and Jan Balaguer and Jackie Xiang and Pidong Wang and Zoe Ashwood and Anton Briukhov and Albert Webson and Sanjay Ganapathy and Smit Sanghavi and Ajay Kannan and Ming-Wei Chang and Axel Stjerngren and Josip Djolonga and Yuting Sun and Ankur Bapna and Matthew Aitchison and Pedram Pejman and Henryk Michalewski and Tianhe Yu and Cindy Wang and Juliette Love and Junwhan Ahn and Dawn Bloxwich and Kehang Han and Peter Humphreys and Thibault Sellam and James Bradbury and Varun Godbole and Sina Samangooei and Bogdan Damoc and Alex Kaskasoli and Sébastien M. R. Arnold and Vijay Vasudevan and Shubham Agrawal and Jason Riesa and Dmitry Lepikhin and Richard Tanburn and Srivatsan Srinivasan and Hyeontaek Lim and Sarah Hodkinson and Pranav Shyam and Johan Ferret and Steven Hand and Ankush Garg and Tom Le Paine and Jian Li and Yujia Li and Minh Giang and Alexander Neitz and Zaheer Abbas and Sarah York and Machel Reid and Elizabeth Cole and Aakanksha Chowdhery and Dipanjan Das and Dominika Rogozińska and Vitaliy Nikolaev and Pablo Sprechmann and Zachary Nado and Lukas Zilka and Flavien Prost and Luheng He and Marianne Monteiro and Gaurav Mishra and Chris Welty and Josh Newlan and Dawei Jia and Miltiadis Allamanis and Clara Huiyi Hu and Raoul de Liedekerke and Justin Gilmer and Carl Saroufim and Shruti Rijhwani and Shaobo Hou and Disha Shrivastava and Anirudh Baddepudi and Alex Goldin and Adnan Ozturel and Albin Cassirer and Yunhan Xu and Daniel Sohn and Devendra Sachan and Reinald Kim Amplayo and Craig Swanson and Dessie Petrova and Shashi Narayan and Arthur Guez and Siddhartha Brahma and Jessica Landon and Miteyan Patel and Ruizhe Zhao and Kevin Villela and Luyu Wang and Wenhao Jia and Matthew Rahtz and Mai Giménez and Legg Yeung and James Keeling and Petko Georgiev and Diana Mincu and Boxi Wu and Salem Haykal and Rachel Saputro and Kiran Vodrahalli and James Qin and Zeynep Cankara and Abhanshu Sharma and Nick Fernando and Will Hawkins and Behnam Neyshabur and Solomon Kim and Adrian Hutter and Priyanka Agrawal and Alex Castro-Ros and George van den Driessche and Tao Wang and Fan Yang and Shuo-yiin Chang and Paul Komarek and Ross McIlroy and Mario Lučić and Guodong Zhang and Wael Farhan and Michael Sharman and Paul Natsev and Paul Michel and Yamini Bansal and Siyuan Qiao and Kris Cao and Siamak Shakeri and Christina Butterfield and Justin Chung and Paul Kishan Rubenstein and Shivani Agrawal and Arthur Mensch and Kedar Soparkar and Karel Lenc and Timothy Chung and Aedan Pope and Loren Maggiore and Jackie Kay and Priya Jhakra and Shibo Wang and Joshua Maynez and Mary Phuong and Taylor Tobin and Andrea Tacchetti and Maja Trebacz and Kevin Robinson and Yash Katariya and Sebastian Riedel and Paige Bailey and Kefan Xiao and Nimesh Ghelani and Lora Aroyo and Ambrose Slone and Neil Houlsby and Xuehan Xiong and Zhen Yang and Elena Gribovskaya and Jonas Adler and Mateo Wirth and Lisa Lee and Music Li and Thais Kagohara and Jay Pavagadhi and Sophie Bridgers and Anna Bortsova and Sanjay Ghemawat and Zafarali Ahmed and Tianqi Liu and Richard Powell and Vijay Bolina and Mariko Iinuma and Polina Zablotskaia and James Besley and Da-Woon Chung and Timothy Dozat and Ramona Comanescu and Xiance Si and Jeremy Greer and Guolong Su and Martin Polacek and Raphaël Lopez Kaufman and Simon Tokumine and Hexiang Hu and Elena Buchatskaya and Yingjie Miao and Mohamed Elhawaty and Aditya Siddhant and Nenad Tomasev and Jinwei Xing and Christina Greer and Helen Miller and Shereen Ashraf and Aurko Roy and Zizhao Zhang and Ada Ma and Angelos Filos and Milos Besta and Rory Blevins and Ted Klimenko and Chih-Kuan Yeh and Soravit Changpinyo and Jiaqi Mu and Oscar Chang and Mantas Pajarskas and Carrie Muir and Vered Cohen and Charline Le Lan and Krishna Haridasan and Amit Marathe and Steven Hansen and Sholto Douglas and Rajkumar Samuel and Mingqiu Wang and Sophia Austin and Chang Lan and Jiepu Jiang and Justin Chiu and Jaime Alonso Lorenzo and Lars Lowe Sjösund and Sébastien Cevey and Zach Gleicher and Thi Avrahami and Anudhyan Boral and Hansa Srinivasan and Vittorio Selo and Rhys May and Konstantinos Aisopos and Léonard Hussenot and Livio Baldini Soares and Kate Baumli and Michael B. Chang and Adrià Recasens and Ben Caine and Alexander Pritzel and Filip Pavetic and Fabio Pardo and Anita Gergely and Justin Frye and Vinay Ramasesh and Dan Horgan and Kartikeya Badola and Nora Kassner and Subhrajit Roy and Ethan Dyer and Víctor Campos Campos and Alex Tomala and Yunhao Tang and Dalia El Badawy and Elspeth White and Basil Mustafa and Oran Lang and Abhishek Jindal and Sharad Vikram and Zhitao Gong and Sergi Caelles and Ross Hemsley and Gregory Thornton and Fangxiaoyu Feng and Wojciech Stokowiec and Ce Zheng and Phoebe Thacker and Çağlar Ünlü and Zhishuai Zhang and Mohammad Saleh and James Svensson and Max Bileschi and Piyush Patil and Ankesh Anand and Roman Ring and Katerina Tsihlas and Arpi Vezer and Marco Selvi and Toby Shevlane and Mikel Rodriguez and Tom Kwiatkowski and Samira Daruki and Keran Rong and Allan Dafoe and Nicholas FitzGerald and Keren Gu-Lemberg and Mina Khan and Lisa Anne Hendricks and Marie Pellat and Vladimir Feinberg and James Cobon-Kerr and Tara Sainath and Maribeth Rauh and Sayed Hadi Hashemi and Richard Ives and Yana Hasson and Eric Noland and Yuan Cao and Nathan Byrd and Le Hou and Qingze Wang and Thibault Sottiaux and Michela Paganini and Jean-Baptiste Lespiau and Alexandre Moufarek and Samer Hassan and Kaushik Shivakumar and Joost van Amersfoort and Amol Mandhane and Pratik Joshi and Anirudh Goyal and Matthew Tung and Andrew Brock and Hannah Sheahan and Vedant Misra and Cheng Li and Nemanja Rakićević and Mostafa Dehghani and Fangyu Liu and Sid Mittal and Junhyuk Oh and Seb Noury and Eren Sezener and Fantine Huot and Matthew Lamm and Nicola De Cao and Charlie Chen and Sidharth Mudgal and Romina Stella and Kevin Brooks and Gautam Vasudevan and Chenxi Liu and Mainak Chain and Nivedita Melinkeri and Aaron Cohen and Venus Wang and Kristie Seymore and Sergey Zubkov and Rahul Goel and Summer Yue and Sai Krishnakumaran and Brian Albert and Nate Hurley and Motoki Sano and Anhad Mohananey and Jonah Joughin and Egor Filonov and Tomasz Kępa and Yomna Eldawy and Jiawern Lim and Rahul Rishi and Shirin Badiezadegan and Taylor Bos and Jerry Chang and Sanil Jain and Sri Gayatri Sundara Padmanabhan and Subha Puttagunta and Kalpesh Krishna and Leslie Baker and Norbert Kalb and Vamsi Bedapudi and Adam Kurzrok and Shuntong Lei and Anthony Yu and Oren Litvin and Xiang Zhou and Zhichun Wu and Sam Sobell and Andrea Siciliano and Alan Papir and Robby Neale and Jonas Bragagnolo and Tej Toor and Tina Chen and Valentin Anklin and Feiran Wang and Richie Feng and Milad Gholami and Kevin Ling and Lijuan Liu and Jules Walter and Hamid Moghaddam and Arun Kishore and Jakub Adamek and Tyler Mercado and Jonathan Mallinson and Siddhinita Wandekar and Stephen Cagle and Eran Ofek and Guillermo Garrido and Clemens Lombriser and Maksim Mukha and Botu Sun and Hafeezul Rahman Mohammad and Josip Matak and Yadi Qian and Vikas Peswani and Pawel Janus and Quan Yuan and Leif Schelin and Oana David and Ankur Garg and Yifan He and Oleksii Duzhyi and Anton Älgmyr and Timothée Lottaz and Qi Li and Vikas Yadav and Luyao Xu and Alex Chinien and Rakesh Shivanna and Aleksandr Chuklin and Josie Li and Carrie Spadine and Travis Wolfe and Kareem Mohamed and Subhabrata Das and Zihang Dai and Kyle He and Daniel von Dincklage and Shyam Upadhyay and Akanksha Maurya and Luyan Chi and Sebastian Krause and Khalid Salama and Pam G Rabinovitch and Pavan Kumar Reddy M and Aarush Selvan and Mikhail Dektiarev and Golnaz Ghiasi and Erdem Guven and Himanshu Gupta and Boyi Liu and Deepak Sharma and Idan Heimlich Shtacher and Shachi Paul and Oscar Akerlund and François-Xavier Aubet and Terry Huang and Chen Zhu and Eric Zhu and Elico Teixeira and Matthew Fritze and Francesco Bertolini and Liana-Eleonora Marinescu and Martin Bölle and Dominik Paulus and Khyatti Gupta and Tejasi Latkar and Max Chang and Jason Sanders and Roopa Wilson and Xuewei Wu and Yi-Xuan Tan and Lam Nguyen Thiet and Tulsee Doshi and Sid Lall and Swaroop Mishra and Wanming Chen and Thang Luong and Seth Benjamin and Jasmine Lee and Ewa Andrejczuk and Dominik Rabiej and Vipul Ranjan and Krzysztof Styrc and Pengcheng Yin and Jon Simon and Malcolm Rose Harriott and Mudit Bansal and Alexei Robsky and Geoff Bacon and David Greene and Daniil Mirylenka and Chen Zhou and Obaid Sarvana and Abhimanyu Goyal and Samuel Andermatt and Patrick Siegler and Ben Horn and Assaf Israel and Francesco Pongetti and Chih-Wei "Louis" Chen and Marco Selvatici and Pedro Silva and Kathie Wang and Jackson Tolins and Kelvin Guu and Roey Yogev and Xiaochen Cai and Alessandro Agostini and Maulik Shah and Hung Nguyen and Noah Ó Donnaile and Sébastien Pereira and Linda Friso and Adam Stambler and Adam Kurzrok and Chenkai Kuang and Yan Romanikhin and Mark Geller and ZJ Yan and Kane Jang and Cheng-Chun Lee and Wojciech Fica and Eric Malmi and Qijun Tan and Dan Banica and Daniel Balle and Ryan Pham and Yanping Huang and Diana Avram and Hongzhi Shi and Jasjot Singh and Chris Hidey and Niharika Ahuja and Pranab Saxena and Dan Dooley and Srividya Pranavi Potharaju and Eileen O'Neill and Anand Gokulchandran and Ryan Foley and Kai Zhao and Mike Dusenberry and Yuan Liu and Pulkit Mehta and Ragha Kotikalapudi and Chalence Safranek-Shrader and Andrew Goodman and Joshua Kessinger and Eran Globen and Prateek Kolhar and Chris Gorgolewski and Ali Ibrahim and Yang Song and Ali Eichenbaum and Thomas Brovelli and Sahitya Potluri and Preethi Lahoti and Cip Baetu and Ali Ghorbani and Charles Chen and Andy Crawford and Shalini Pal and Mukund Sridhar and Petru Gurita and Asier Mujika and Igor Petrovski and Pierre-Louis Cedoz and Chenmei Li and Shiyuan Chen and Niccolò Dal Santo and Siddharth Goyal and Jitesh Punjabi and Karthik Kappaganthu and Chester Kwak and Pallavi LV and Sarmishta Velury and Himadri Choudhury and Jamie Hall and Premal Shah and Ricardo Figueira and Matt Thomas and Minjie Lu and Ting Zhou and Chintu Kumar and Thomas Jurdi and Sharat Chikkerur and Yenai Ma and Adams Yu and Soo Kwak and Victor Ähdel and Sujeevan Rajayogam and Travis Choma and Fei Liu and Aditya Barua and Colin Ji and Ji Ho Park and Vincent Hellendoorn and Alex Bailey and Taylan Bilal and Huanjie Zhou and Mehrdad Khatir and Charles Sutton and Wojciech Rzadkowski and Fiona Macintosh and Roopali Vij and Konstantin Shagin and Paul Medina and Chen Liang and Jinjing Zhou and Pararth Shah and Yingying Bi and Attila Dankovics and Shipra Banga and Sabine Lehmann and Marissa Bredesen and Zifan Lin and John Eric Hoffmann and Jonathan Lai and Raynald Chung and Kai Yang and Nihal Balani and Arthur Bražinskas and Andrei Sozanschi and Matthew Hayes and Héctor Fernández Alcalde and Peter Makarov and Will Chen and Antonio Stella and Liselotte Snijders and Michael Mandl and Ante Kärrman and Paweł Nowak and Xinyi Wu and Alex Dyck and Krishnan Vaidyanathan and Raghavender R and Jessica Mallet and Mitch Rudominer and Eric Johnston and Sushil Mittal and Akhil Udathu and Janara Christensen and Vishal Verma and Zach Irving and Andreas Santucci and Gamaleldin Elsayed and Elnaz Davoodi and Marin Georgiev and Ian Tenney and Nan Hua and Geoffrey Cideron and Edouard Leurent and Mahmoud Alnahlawi and Ionut Georgescu and Nan Wei and Ivy Zheng and Dylan Scandinaro and Heinrich Jiang and Jasper Snoek and Mukund Sundararajan and Xuezhi Wang and Zack Ontiveros and Itay Karo and Jeremy Cole and Vinu Rajashekhar and Lara Tumeh and Eyal Ben-David and Rishub Jain and Jonathan Uesato and Romina Datta and Oskar Bunyan and Shimu Wu and John Zhang and Piotr Stanczyk and Ye Zhang and David Steiner and Subhajit Naskar and Michael Azzam and Matthew Johnson and Adam Paszke and Chung-Cheng Chiu and Jaume Sanchez Elias and Afroz Mohiuddin and Faizan Muhammad and Jin Miao and Andrew Lee and Nino Vieillard and Jane Park and Jiageng Zhang and Jeff Stanway and Drew Garmon and Abhijit Karmarkar and Zhe Dong and Jong Lee and Aviral Kumar and Luowei Zhou and Jonathan Evens and William Isaac and Geoffrey Irving and Edward Loper and Michael Fink and Isha Arkatkar and Nanxin Chen and Izhak Shafran and Ivan Petrychenko and Zhe Chen and Johnson Jia and Anselm Levskaya and Zhenkai Zhu and Peter Grabowski and Yu Mao and Alberto Magni and Kaisheng Yao and Javier Snaider and Norman Casagrande and Evan Palmer and Paul Suganthan and Alfonso Castaño and Irene Giannoumis and Wooyeol Kim and Mikołaj Rybiński and Ashwin Sreevatsa and Jennifer Prendki and David Soergel and Adrian Goedeckemeyer and Willi Gierke and Mohsen Jafari and Meenu Gaba and Jeremy Wiesner and Diana Gage Wright and Yawen Wei and Harsha Vashisht and Yana Kulizhskaya and Jay Hoover and Maigo Le and Lu Li and Chimezie Iwuanyanwu and Lu Liu and Kevin Ramirez and Andrey Khorlin and Albert Cui and Tian LIN and Marcus Wu and Ricardo Aguilar and Keith Pallo and Abhishek Chakladar and Ginger Perng and Elena Allica Abellan and Mingyang Zhang and Ishita Dasgupta and Nate Kushman and Ivo Penchev and Alena Repina and Xihui Wu and Tom van der Weide and Priya Ponnapalli and Caroline Kaplan and Jiri Simsa and Shuangfeng Li and Olivier Dousse and Fan Yang and Jeff Piper and Nathan Ie and Rama Pasumarthi and Nathan Lintz and Anitha Vijayakumar and Daniel Andor and Pedro Valenzuela and Minnie Lui and Cosmin Paduraru and Daiyi Peng and Katherine Lee and Shuyuan Zhang and Somer Greene and Duc Dung Nguyen and Paula Kurylowicz and Cassidy Hardin and Lucas Dixon and Lili Janzer and Kiam Choo and Ziqiang Feng and Biao Zhang and Achintya Singhal and Dayou Du and Dan McKinnon and Natasha Antropova and Tolga Bolukbasi and Orgad Keller and David Reid and Daniel Finchelstein and Maria Abi Raad and Remi Crocker and Peter Hawkins and Robert Dadashi and Colin Gaffney and Ken Franko and Anna Bulanova and Rémi Leblond and Shirley Chung and Harry Askham and Luis C. Cobo and Kelvin Xu and Felix Fischer and Jun Xu and Christina Sorokin and Chris Alberti and Chu-Cheng Lin and Colin Evans and Alek Dimitriev and Hannah Forbes and Dylan Banarse and Zora Tung and Mark Omernick and Colton Bishop and Rachel Sterneck and Rohan Jain and Jiawei Xia and Ehsan Amid and Francesco Piccinno and Xingyu Wang and Praseem Banzal and Daniel J. Mankowitz and Alex Polozov and Victoria Krakovna and Sasha Brown and MohammadHossein Bateni and Dennis Duan and Vlad Firoiu and Meghana Thotakuri and Tom Natan and Matthieu Geist and Ser tan Girgin and Hui Li and Jiayu Ye and Ofir Roval and Reiko Tojo and Michael Kwong and James Lee-Thorp and Christopher Yew and Danila Sinopalnikov and Sabela Ramos and John Mellor and Abhishek Sharma and Kathy Wu and David Miller and Nicolas Sonnerat and Denis Vnukov and Rory Greig and Jennifer Beattie and Emily Caveness and Libin Bai and Julian Eisenschlos and Alex Korchemniy and Tomy Tsai and Mimi Jasarevic and Weize Kong and Phuong Dao and Zeyu Zheng and Frederick Liu and Fan Yang and Rui Zhu and Tian Huey Teh and Jason Sanmiya and Evgeny Gladchenko and Nejc Trdin and Daniel Toyama and Evan Rosen and Sasan Tavakkol and Linting Xue and Chen Elkind and Oliver Woodman and John Carpenter and George Papamakarios and Rupert Kemp and Sushant Kafle and Tanya Grunina and Rishika Sinha and Alice Talbert and Diane Wu and Denese Owusu-Afriyie and Cosmo Du and Chloe Thornton and Jordi Pont-Tuset and Pradyumna Narayana and Jing Li and Saaber Fatehi and John Wieting and Omar Ajmeri and Benigno Uria and Yeongil Ko and Laura Knight and Amélie Héliou and Ning Niu and Shane Gu and Chenxi Pang and Yeqing Li and Nir Levine and Ariel Stolovich and Rebeca Santamaria-Fernandez and Sonam Goenka and Wenny Yustalim and Robin Strudel and Ali Elqursh and Charlie Deck and Hyo Lee and Zonglin Li and Kyle Levin and Raphael Hoffmann and Dan Holtmann-Rice and Olivier Bachem and Sho Arora and Christy Koh and Soheil Hassas Yeganeh and Siim Põder and Mukarram Tariq and Yanhua Sun and Lucian Ionita and Mojtaba Seyedhosseini and Pouya Tafti and Zhiyu Liu and Anmol Gulati and Jasmine Liu and Xinyu Ye and Bart Chrzaszcz and Lily Wang and Nikhil Sethi and Tianrun Li and Ben Brown and Shreya Singh and Wei Fan and Aaron Parisi and Joe Stanton and Vinod Koverkathu and Christopher A. Choquette-Choo and Yunjie Li and TJ Lu and Abe Ittycheriah and Prakash Shroff and Mani Varadarajan and Sanaz Bahargam and Rob Willoughby and David Gaddy and Guillaume Desjardins and Marco Cornero and Brona Robenek and Bhavishya Mittal and Ben Albrecht and Ashish Shenoy and Fedor Moiseev and Henrik Jacobsson and Alireza Ghaffarkhah and Morgane Rivière and Alanna Walton and Clément Crepy and Alicia Parrish and Zongwei Zhou and Clement Farabet and Carey Radebaugh and Praveen Srinivasan and Claudia van der Salm and Andreas Fidjeland and Salvatore Scellato and Eri Latorre-Chimoto and Hanna Klimczak-Plucińska and David Bridson and Dario de Cesare and Tom Hudson and Piermaria Mendolicchio and Lexi Walker and Alex Morris and Matthew Mauger and Alexey Guseynov and Alison Reid and Seth Odoom and Lucia Loher and Victor Cotruta and Madhavi Yenugula and Dominik Grewe and Anastasia Petrushkina and Tom Duerig and Antonio Sanchez and Steve Yadlowsky and Amy Shen and Amir Globerson and Lynette Webb and Sahil Dua and Dong Li and Surya Bhupatiraju and Dan Hurt and Haroon Qureshi and Ananth Agarwal and Tomer Shani and Matan Eyal and Anuj Khare and Shreyas Rammohan Belle and Lei Wang and Chetan Tekur and Mihir Sanjay Kale and Jinliang Wei and Ruoxin Sang and Brennan Saeta and Tyler Liechty and Yi Sun and Yao Zhao and Stephan Lee and Pandu Nayak and Doug Fritz and Manish Reddy Vuyyuru and John Aslanides and Nidhi Vyas and Martin Wicke and Xiao Ma and Evgenii Eltyshev and Nina Martin and Hardie Cate and James Manyika and Keyvan Amiri and Yelin Kim and Xi Xiong and Kai Kang and Florian Luisier and Nilesh Tripuraneni and David Madras and Mandy Guo and Austin Waters and Oliver Wang and Joshua Ainslie and Jason Baldridge and Han Zhang and Garima Pruthi and Jakob Bauer and Feng Yang and Riham Mansour and Jason Gelman and Yang Xu and George Polovets and Ji Liu and Honglong Cai and Warren Chen and XiangHai Sheng and Emily Xue and Sherjil Ozair and Christof Angermueller and Xiaowei Li and Anoop Sinha and Weiren Wang and Julia Wiesinger and Emmanouil Koukoumidis and Yuan Tian and Anand Iyer and Madhu Gurumurthy and Mark Goldenson and Parashar Shah and MK Blake and Hongkun Yu and Anthony Urbanowicz and Jennimaria Palomaki and Chrisantha Fernando and Ken Durden and Harsh Mehta and Nikola Momchev and Elahe Rahimtoroghi and Maria Georgaki and Amit Raul and Sebastian Ruder and Morgan Redshaw and Jinhyuk Lee and Denny Zhou and Komal Jalan and Dinghua Li and Blake Hechtman and Parker Schuh and Milad Nasr and Kieran Milan and Vladimir Mikulik and Juliana Franco and Tim Green and Nam Nguyen and Joe Kelley and Aroma Mahendru and Andrea Hu and Joshua Howland and Ben Vargas and Jeffrey Hui and Kshitij Bansal and Vikram Rao and Rakesh Ghiya and Emma Wang and Ke Ye and Jean Michel Sarr and Melanie Moranski Preston and Madeleine Elish and Steve Li and Aakash Kaku and Jigar Gupta and Ice Pasupat and Da-Cheng Juan and Milan Someswar and Tejvi M. and Xinyun Chen and Aida Amini and Alex Fabrikant and Eric Chu and Xuanyi Dong and Amruta Muthal and Senaka Buthpitiya and Sarthak Jauhari and Nan Hua and Urvashi Khandelwal and Ayal Hitron and Jie Ren and Larissa Rinaldi and Shahar Drath and Avigail Dabush and Nan-Jiang Jiang and Harshal Godhia and Uli Sachs and Anthony Chen and Yicheng Fan and Hagai Taitelbaum and Hila Noga and Zhuyun Dai and James Wang and Chen Liang and Jenny Hamer and Chun-Sung Ferng and Chenel Elkind and Aviel Atias and Paulina Lee and Vít Listík and Mathias Carlen and Jan van de Kerkhof and Marcin Pikus and Krunoslav Zaher and Paul Müller and Sasha Zykova and Richard Stefanec and Vitaly Gatsko and Christoph Hirnschall and Ashwin Sethi and Xingyu Federico Xu and Chetan Ahuja and Beth Tsai and Anca Stefanoiu and Bo Feng and Keshav Dhandhania and Manish Katyal and Akshay Gupta and Atharva Parulekar and Divya Pitta and Jing Zhao and Vivaan Bhatia and Yashodha Bhavnani and Omar Alhadlaq and Xiaolin Li and Peter Danenberg and Dennis Tu and Alex Pine and Vera Filippova and Abhipso Ghosh and Ben Limonchik and Bhargava Urala and Chaitanya Krishna Lanka and Derik Clive and Yi Sun and Edward Li and Hao Wu and Kevin Hongtongsak and Ianna Li and Kalind Thakkar and Kuanysh Omarov and Kushal Majmundar and Michael Alverson and Michael Kucharski and Mohak Patel and Mudit Jain and Maksim Zabelin and Paolo Pelagatti and Rohan Kohli and Saurabh Kumar and Joseph Kim and Swetha Sankar and Vineet Shah and Lakshmi Ramachandruni and Xiangkai Zeng and Ben Bariach and Laura Weidinger and Tu Vu and Alek Andreev and Antoine He and Kevin Hui and Sheleem Kashem and Amar Subramanya and Sissie Hsiao and Demis Hassabis and Koray Kavukcuoglu and Adam Sadovsky and Quoc Le and Trevor Strohman and Yonghui Wu and Slav Petrov and Jeffrey Dean and Oriol Vinyals},
      year={2025},
      eprint={2312.11805},
      archivePrefix={arXiv},
      primaryClass={cs.CL},
      url={https://arxiv.org/abs/2312.11805}, 
      accessed={9 March 2026}
}

@inproceedings{vaswani2017,
 author = {Vaswani, Ashish and Shazeer, Noam and Parmar, Niki and Uszkoreit, Jakob and Jones, Llion and Gomez, Aidan N and Kaiser, \L ukasz and Polosukhin, Illia},
 booktitle = {Advances in Neural Information Processing Systems},
 editor = {I. Guyon and U. Von Luxburg and S. Bengio and H. Wallach and R. Fergus and S. Vishwanathan and R. Garnett},
 pages = {},
 publisher = {Curran Associates, Inc.},
 title = {Attention is All you Need},
 url = {https://proceedings.neurips.cc/paper\_files/paper/2017/file/3f5ee243547dee91fbd053c1c4a845aa-Paper.pdf},
 volume = {30},
 year = {2017}
}

@misc{zeller2026,
      title={MentisOculi: Revealing the Limits of Reasoning with Mental Imagery}, 
      author={Jana Zeller and Thaddäus Wiedemer and Fanfei Li and Thomas Klein and Prasanna Mayilvahanan and Matthias Bethge and Felix Wichmann and Ryan Cotterell and Wieland Brendel},
      year={2026},
      eprint={2602.02465},
      archivePrefix={arXiv},
      primaryClass={cs.AI},
      url={https://arxiv.org/abs/2602.02465}, 
      accessed={9 March 2026}
}

@misc{plunkett2025,
      title={Self-Interpretability: LLMs Can Describe Complex Internal Processes that Drive Their Decisions, and Improve with Training}, 
      author={Dillon Plunkett and Adam Morris and Keerthi Reddy and Jorge Morales},
      year={2025},
      eprint={2505.17120},
      archivePrefix={arXiv},
      primaryClass={cs.CL},
      url={https://arxiv.org/abs/2505.17120}, 
      accessed={9 March 2026}
}

@misc{yang2025,
      title={Machine Mental Imagery: Empower Multimodal Reasoning with Latent Visual Tokens}, 
      author={Zeyuan Yang and Xueyang Yu and Delin Chen and Maohao Shen and Chuang Gan},
      year={2025},
      eprint={2506.17218},
      archivePrefix={arXiv},
      primaryClass={cs.CV},
      url={https://arxiv.org/abs/2506.17218}, 
      accessed={9 March 2026}
}

@misc{sepehri2026,
      title={Hyperphantasia: A Benchmark for Evaluating the Mental Visualization Capabilities of Multimodal LLMs}, 
      author={Mohammad Shahab Sepehri and Berk Tinaz and Zalan Fabian and Mahdi Soltanolkotabi},
      year={2026},
      eprint={2507.11932},
      archivePrefix={arXiv},
      primaryClass={cs.CV},
      url={https://arxiv.org/abs/2507.11932},
      accessed={9 March 2026}
}

@inproceedings{
prakash2026,
title={Language Models Use Lookbacks to Track Beliefs},
author={Nikhil Prakash and Natalie Shapira and Arnab Sen Sharma and Christoph Riedl and Yonatan Belinkov and Tamar Rott Shaham and David Bau and Atticus Geiger},
booktitle={The Fourteenth International Conference on Learning Representations},
year={2026},
url={https://openreview.net/forum?id=6gO6KTRMpG},
}

@article{
kosinski2024,
author = {Michal Kosinski },
title = {Evaluating large language models in theory of mind tasks},
journal = {Proceedings of the National Academy of Sciences},
volume = {121},
number = {45},
pages = {e2405460121},
year = {2024},
doi = {10.1073/pnas.2405460121},
URL = {https://www.pnas.org/doi/abs/10.1073/pnas.2405460121},
eprint = {https://www.pnas.org/doi/pdf/10.1073/pnas.2405460121}}

@article{hu2023,
    author = {Hu, Jennifer and Sosa, Felix and Ullman, Tomer},
    title = {Re-evaluating Theory of Mind evaluation in large language models},
    journal = {Philosophical Transactions of the Royal Society B: Biological Sciences},
    volume = {380},
    number = {1932},
    pages = {20230499},
    year = {2025},
    month = {08},
    issn = {0962-8436},
    doi = {10.1098/rstb.2023.0499},
    url = {https://doi.org/10.1098/rstb.2023.0499},
    eprint = {https://royalsocietypublishing.org/rstb/article-pdf/doi/10.1098/rstb.2023.0499/2821303/rstb.2023.0499.pdf},
}

@inproceedings{
ackerman2026,
title={Evidence for Limited Metacognition in {LLM}s},
author={Christopher Ackerman},
booktitle={The Fourteenth International Conference on Learning Representations},
year={2026},
url={https://openreview.net/forum?id=gb9HR8hxtU}
}

@misc{lindsey2026,
      title={Emergent Introspective Awareness in Large Language Models}, 
      author={Jack Lindsey},
      year={2026},
      eprint={2601.01828},
      archivePrefix={arXiv},
      primaryClass={cs.CL},
      url={https://arxiv.org/abs/2601.01828}, 
      accessed={9 March 2026}
}

@inproceedings{
binder2025,
title={Looking Inward: Language Models Can Learn About Themselves by Introspection},
author={Felix Jedidja Binder and James Chua and Tomek Korbak and Henry Sleight and John Hughes and Robert Long and Ethan Perez and Miles Turpin and Owain Evans},
booktitle={The Thirteenth International Conference on Learning Representations},
year={2025},
url={https://openreview.net/forum?id=eb5pkwIB5i},
}

@misc{hendrycks2021,
      title={Measuring Massive Multitask Language Understanding}, 
      author={Dan Hendrycks and Collin Burns and Steven Basart and Andy Zou and Mantas Mazeika and Dawn Song and Jacob Steinhardt},
      year={2021},
      eprint={2009.03300},
      archivePrefix={arXiv},
      primaryClass={cs.CY},
      url={https://arxiv.org/abs/2009.03300}, 
}

@article{kocisky2018,
    title = {The {N}arrative{QA} Reading Comprehension Challenge},
    author = {Ko{\v{c}}isk{\'y}, Tom{\'a}{\v{s}}  and
      Schwarz, Jonathan  and
      Blunsom, Phil  and
      Dyer, Chris  and
      Hermann, Karl Moritz  and
      Melis, G{\'a}bor  and
      Grefenstette, Edward",
    editor = "Lee, Lillian  and
      Johnson, Mark  and
      Toutanova, Kristina  and
      Roark, Brian},
    journal = {Transactions of the Association for Computational Linguistics},
    volume = {6},
    year = {2018},
    address = {Cambridge, MA},
    publisher = {MIT Press},
    url = {https://aclanthology.org/Q18-1023/},
    doi = {10.1162/tacl\_a\_00023},
    pages = {317--328}
}

@misc{rein2023,
      title={GPQA: A Graduate-Level Google-Proof Q\&A Benchmark}, 
      author={David Rein and Betty Li Hou and Asa Cooper Stickland and Jackson Petty and Richard Yuanzhe Pang and Julien Dirani and Julian Michael and Samuel R. Bowman},
      year={2023},
      eprint={2311.12022},
      archivePrefix={arXiv},
      primaryClass={cs.AI},
      url={https://arxiv.org/abs/2311.12022}, 
}

@article{finke1989,
  title={Reinterpreting visual patterns in mental imagery},
  author={Finke, Ronald A and Pinker, Steven and Farah, Martha J},
  journal={Cognitive Science},
  volume={13},
  number={1},
  pages={51--78},
  year={1989},
  publisher={Elsevier}
}

@article{dance2022,
title = {The prevalence of aphantasia (imagery weakness) in the general population},
journal = {Consciousness and Cognition},
volume = {97},
pages = {103243},
year = {2022},
issn = {1053-8100},
doi = {https://doi.org/10.1016/j.concog.2021.103243},
url = {https://www.sciencedirect.com/science/article/pii/S1053810021001690},
author = {C.J. Dance and A. Ipser and J. Simner}
}

@article{bocchi2016,
title = {The Key of the Maze: The role of mental imagery and cognitive flexibility in navigational planning},
journal = {Neuroscience Letters},
volume = {651},
pages = {146-150},
year = {2017},
issn = {0304-3940},
doi = {https://doi.org/10.1016/j.neulet.2017.05.009},
url = {https://www.sciencedirect.com/science/article/pii/S0304394017303877},
author = {Alessia Bocchi and Marika Carrieri and Stefania Lancia and Valentina Quaresima and Laura Piccardi}
}

@article{farrington2011,
author = {Farrington, Jeanne},
title = {Seven plus or minus two},
journal = {Performance Improvement Quarterly},
volume = {23},
number = {4},
pages = {113-116},
doi = {https://doi.org/10.1002/piq.20099},
url = {https://onlinelibrary.wiley.com/doi/abs/10.1002/piq.20099},
eprint = {https://onlinelibrary.wiley.com/doi/pdf/10.1002/piq.20099},
year = {2011}
}

@misc{wurgaft2025,
      title={In-Context Learning Strategies Emerge Rationally}, 
      author={Daniel Wurgaft and Ekdeep Singh Lubana and Core Francisco Park and Hidenori Tanaka and Gautam Reddy and Noah D. Goodman},
      year={2025},
      eprint={2506.17859},
      archivePrefix={arXiv},
      primaryClass={cs.LG},
      url={https://arxiv.org/abs/2506.17859}, 
}

@article{miller1956,
  title={The magical number seven, plus or minus two: Some limits on our capacity for processing information.},
  author={Miller, George A.},
  journal={Psychological Review},
  volume={63},
  number={2},
  pages={81-97},
  year={1956},
doi= {10.1037/h0043158}
}

@article{pounder2022,
  title={Only minimal differences between individuals with congenital aphantasia and those with typical imagery on neuropsychological tasks that involve imagery},
  author={Zoë Pounder and Jane Jacob and Samuel Evans and Catherine Loveday and Alison F. Eardley and Juha Silvanto},
  journal={Cortex},
  volume={148},
  pages={180-192},
  year={2022},
doi={10.1037/h0043158}
}

@article{kay2024,
title = {Slower but more accurate mental rotation performance in aphantasia linked to differences in cognitive strategies},
journal = {Consciousness and Cognition},
volume = {121},
pages = {103694},
year = {2024},
issn = {1053-8100},
doi = {https://doi.org/10.1016/j.concog.2024.103694},
url = {https://www.sciencedirect.com/science/article/pii/S1053810024000618},
author = {Lachlan Kay and Rebecca Keogh and Joel Pearson}
}

@article{pylyshyn2002, title={Mental imagery: In search of a theory}, volume={25}, DOI={10.1017/S0140525X02000043}, number={2}, journal={Behavioral and Brain Sciences}, author={Pylyshyn, Zenon W.}, year={2002}, pages={157–182}}

@article{pylyshyn1973,
  title={What the mind's eye tells the mind's brain: A critique of mental imagery.},
  author={Pylyshyn, Zenon W},
  journal={Psychological bulletin},
  volume={80},
  number={1},
  pages={1},
  year={1973},
  publisher={American Psychological Association}
}

@article{
shepard1971,
author = {Roger N. Shepard  and Jacqueline Metzler },
title = {Mental Rotation of Three-Dimensional Objects},
journal = {Science},
volume = {171},
number = {3972},
pages = {701-703},
year = {1971},
doi = {10.1126/science.171.3972.701},
URL = {https://www.science.org/doi/abs/10.1126/science.171.3972.701},
eprint = {https://www.science.org/doi/pdf/10.1126/science.171.3972.701}}

@article{palombo2018,
  title={Individual differences in autobiographical memory},
  author={Palombo, Daniela J and Sheldon, Signy and Levine, Brian},
  journal={Trends in Cognitive Sciences},
  volume={22},
  number={7},
  pages={583--597},
  year={2018},
  publisher={Elsevier}
}

@article{bigelow2023,
  title={Non-commitment in mental imagery},
  author={Bigelow, Eric J. and John P. McCoy and Tomer D. Ullman},
  journal={Cognition},
  volume={238},
  pages={105498},
  year={2023}
}

@article{
wheeler2000,
author = {Mark E. Wheeler  and Steven E. Petersen  and Randy L. Buckner },
title = {Memory's echo: Vivid remembering reactivates
 sensory-specific cortex},
journal = {Proceedings of the National Academy of Sciences},
volume = {97},
number = {20},
pages = {11125-11129},
year = {2000},
doi = {10.1073/pnas.97.20.11125},
URL = {https://www.pnas.org/doi/abs/10.1073/pnas.97.20.11125},
eprint = {https://www.pnas.org/doi/pdf/10.1073/pnas.97.20.11125}}

@article{byrne2007,
  title={Remembering the past and imagining the future: a neural model of spatial memory and imagery.},
  author={Byrne, Patrick and Becker, Suzanna and Burgess, Neil},
  journal={Psychological review},
  volume={114},
  number={2},
  pages={340},
  year={2007},
  publisher={American Psychological Association}
}

@article{holmes2010,
  title={Mental imagery in emotion and emotional disorders},
  author={Holmes, Emily A and Mathews, Andrew},
  journal={Clinical psychology review},
  volume={30},
  number={3},
  pages={349--362},
  year={2010},
  publisher={Elsevier}
}

@article{krasich2024,
author = {Krasich, Kristina and O'Neill, Kevin and De Brigard, Felipe},
title = {Looking at Mental Images: Eye-Tracking Mental Simulation During Retrospective Causal Judgment},
journal = {Cognitive Science},
volume = {48},
number = {3},
pages = {e13426},
doi = {https://doi.org/10.1111/cogs.13426},
url = {https://onlinelibrary.wiley.com/doi/abs/10.1111/cogs.13426},
eprint = {https://onlinelibrary.wiley.com/doi/pdf/10.1111/cogs.13426},
year = {2024}
}

@article{zeman2015,
title = {Lives without imagery – Congenital aphantasia},
journal = {Cortex},
volume = {73},
pages = {378-380},
year = {2015},
issn = {0010-9452},
doi = {https://doi.org/10.1016/j.cortex.2015.05.019},
url = {https://www.sciencedirect.com/science/article/pii/S0010945215001781},
author = {Adam Zeman and Michaela Dewar and Sergio {Della Sala}}
}

@article{zeman2024,
  title={Aphantasia and hyperphantasia: exploring imagery vividness extremes},
  author={Zeman, Adam},
  journal={Trends in Cognitive Sciences},
  volume={28},
  number={5},
  pages={467--480},
  year={2024},
  publisher={Elsevier}
}

@article{marks1973,
author = {Marks, David F.},
title = {VISUAL IMAGERY DIFFERENCES IN THE RECALL OF PICTURES},
journal = {British Journal of Psychology},
volume = {64},
number = {1},
pages = {17-24},
doi = {https://doi.org/10.1111/j.2044-8295.1973.tb01322.x},
url = {https://bpspsychub.onlinelibrary.wiley.com/doi/abs/10.1111/j.2044-8295.1973.tb01322.x},
eprint = {https://bpspsychub.onlinelibrary.wiley.com/doi/pdf/10.1111/j.2044-8295.1973.tb01322.x},
year = {1973}
}

@inproceedings{wang2024,
 author = {Wang, Yubo and Ma, Xueguang and Zhang, Ge and Ni, Yuansheng and Chandra, Abhranil and Guo, Shiguang and Ren, Weiming and Arulraj, Aaran and He, Xuan and Jiang, Ziyan and Li, Tianle and Ku, Max and Wang, Kai and Zhuang, Alex and Fan, Rongqi and Yue, Xiang and Chen, Wenhu},
 booktitle = {Advances in Neural Information Processing Systems},
 editor = {A. Globerson and L. Mackey and D. Belgrave and A. Fan and U. Paquet and J. Tomczak and C. Zhang},
 pages = {95266--95290},
 publisher = {Curran Associates, Inc.},
 title = {MMLU-Pro: A More Robust and Challenging Multi-Task Language Understanding Benchmark},
 url = {https://proceedings.neurips.cc/paper\_files/paper/2024/file/ad236edc564f3e3156e1b2feafb99a24-Paper-Datasets\_and\_Benchmarks\_Track.pdf},
 volume = {37},
 year = {2024}
}

@misc{khan2023,
      title={xCodeEval: A Large Scale Multilingual Multitask Benchmark for Code Understanding, Generation, Translation and Retrieval}, 
      author={Mohammad Abdullah Matin Khan and M Saiful Bari and Xuan Long Do and Weishi Wang and Md Rizwan Parvez and Shafiq Joty},
      year={2023},
      eprint={2303.03004},
      archivePrefix={arXiv},
      primaryClass={cs.CL},
      url={https://arxiv.org/abs/2303.03004}, 
}

@inproceedings{wu2024,
 author = {Wu, Wenshan and Mao, Shaoguang and Zhang, Yadong and Xia, Yan and Dong, Li and Cui, Lei and Wei, Furu},
 booktitle = {Advances in Neural Information Processing Systems},
 editor = {A. Globerson and L. Mackey and D. Belgrave and A. Fan and U. Paquet and J. Tomczak and C. Zhang},
 pages = {90277--90317},
 publisher = {Curran Associates, Inc.},
 title = {Mind\textquotesingle s Eye of LLMs: Visualization-of-Thought Elicits Spatial Reasoning in Large Language Models},
 url = {https://proceedings.neurips.cc/paper\_files/paper/2024/file/a45296e83b19f656392e0130d9e53cb1-Paper-Conference.pdf},
 volume = {37},
 year = {2024}
}

@article{mcgrath2024,
author = {Sam Whitman McGrath and Jacob Russin and Ellie Pavlick and Roman Feiman},
title ={How Can Deep Neural Networks Inform Theory in Psychological Science?},

journal = {Current Directions in Psychological Science},
volume = {33},
number = {5},
pages = {325-333},
year = {2024},
doi = {10.1177/09637214241268098},

URL = { 
    
        https://doi.org/10.1177/09637214241268098
    
    

},
eprint = { 
    
        https://doi.org/10.1177/09637214241268098
    
    

}
}

@inproceedings{leshinskaya2025,
  title={Cognitively Inspired Interpretability in Large Neural Networks},
  author={Leshinskaya, Anna and Webb, Taylor and Pavlick, Ellie and Feng, Jiahai and Opielka, Gustaw and Stevenson, Claire and Blank, Idan A},
  booktitle={Proceedings of the Annual Meeting of the Cognitive Science Society},
  volume={47},
  year={2025}
}

@article{
chen2025,
author = {Zirui Chen  and Michael F. Bonner },
title = {Universal dimensions of visual representation},
journal = {Science Advances},
volume = {11},
number = {27},
pages = {eadw7697},
year = {2025},
doi = {10.1126/sciadv.adw7697},
URL = {https://www.science.org/doi/abs/10.1126/sciadv.adw7697},
eprint = {https://www.science.org/doi/pdf/10.1126/sciadv.adw7697}}

@article{
yamins2014,
author = {Daniel L. K. Yamins  and Ha Hong  and Charles F. Cadieu  and Ethan A. Solomon  and Darren Seibert  and James J. DiCarlo },
title = {Performance-optimized hierarchical models predict neural responses in higher visual cortex},
journal = {Proceedings of the National Academy of Sciences},
volume = {111},
number = {23},
pages = {8619-8624},
year = {2014},
doi = {10.1073/pnas.1403112111},
URL = {https://www.pnas.org/doi/abs/10.1073/pnas.1403112111},
eprint = {https://www.pnas.org/doi/pdf/10.1073/pnas.1403112111}}

@article{faw2009,
author = {Faw, Bill},
year = {2009},
month = {01},
pages = {45-68},
title = {Conflicting Intuitions May Be Based On Differing Abilities: Evidence from Mental Imaging Research},
volume = {16},
journal = {Journal of Consciousness Studies}
}

@article{nanay2021,
author = {Nanay, Bence },
title = {Unconscious mental imagery},
journal = {Philosophical Transactions of the Royal Society B: Biological Sciences},
volume = {376},
number = {1817},
pages = {20190689},
year = {2021},
doi = {10.1098/rstb.2019.0689},

URL = {https://royalsocietypublishing.org/doi/abs/10.1098/rstb.2019.0689},
eprint = {https://royalsocietypublishing.org/doi/pdf/10.1098/rstb.2019.0689}
}

@article{michel2025,
    author = {Michel, Matthias and Morales, Jorge and Block, Ned and Lau, Hakwan},
    title = {Aphantasia as imagery blindsight},
    journal = {Trends in Cognitive Sciences},
    year = {2025},
    doi = {10.1016/j.tics.2024.11.002},
    pages = {8-9},
    volume = {29},
    number = {1}
}

@article{keogh2021,
title = {Visual working memory in aphantasia: Retained accuracy and capacity with a different strategy},
journal = {Cortex},
volume = {143},
pages = {237-253},
year = {2021},
issn = {0010-9452},
doi = {https://doi.org/10.1016/j.cortex.2021.07.012},
url = {https://www.sciencedirect.com/science/article/pii/S0010945221002628},
author = {Rebecca Keogh and Marcus Wicken and Joel Pearson}
}

@article{blomkvist2023,
author = {Blomkvist, Andrea},
title = {Aphantasia: In search of a theory},
journal = {Mind \& Language},
volume = {38},
number = {3},
pages = {866-888},
doi = {https://doi.org/10.1111/mila.12432},
url = {https://onlinelibrary.wiley.com/doi/abs/10.1111/mila.12432},
eprint = {https://onlinelibrary.wiley.com/doi/pdf/10.1111/mila.12432},
year = {2023}
}

\end{document}